%% file: main.tex
\documentclass[12pt]{article}
\usepackage[utf8]{inputenc}

\usepackage{amsfonts,amsmath}
\usepackage{hyperref}
\usepackage{graphicx}
\usepackage{float}
\usepackage[caption=false]{subfig}
\usepackage{booktabs}
\usepackage{multirow}
\usepackage{algorithmic}
\usepackage[titlenumbered,ruled]{algorithm2e}
\usepackage[square,numbers]{natbib}
\title{Fully Bayesian inference for latent variable Gaussian process models}
\author{
Suraj Yerramilli\thanks{Department of Industrial Engineering \& Management Sciences, Northwestern University} \and
Akshay Iyer \thanks{Department of Mechanical Engineering, Northwestern University}\and
Wei Chen \thanks{Department of Mechanical Engineering, Northwestern University} \and
Daniel W. Apley\thanks{Department of Industrial Engineering \& Management Sciences, Northwestern University }
}
\date{}

\newcommand{\smb}[1]{\left(#1\right)}
\newcommand{\bSigma}{\boldsymbol{\Sigma}}
\newcommand{\btheta}{\boldsymbol{\theta}}

\newcommand{\sumObs}[2]{\sum_{#1=1}^{#2}}
\newcommand{\Exp}[1]{\mathbb{E}\left[#1\right]}
\newcommand{\Var}[1]{\text{Var}\left[#1\right]}
\begin{document}

\maketitle

\begin{abstract}
Real engineering and scientific applications often involve one or more qualitative inputs. Standard Gaussian processes (GPs), however, cannot directly accommodate qualitative inputs. The recently introduced latent variable Gaussian process (LVGP) overcomes this issue by first mapping each qualitative factor to underlying latent variables (LVs), and then uses any standard GP covariance function over these LVs. The LVs are estimated similarly to the other GP hyperparameters through maximum likelihood estimation, and then plugged into the prediction expressions. However, this plug-in approach will not account for uncertainty in estimation of the LVs, which can be significant especially with limited training data. In this work, we develop a fully Bayesian approach for the LVGP model and for visualizing the effects of the qualitative inputs via their LVs. We also develop approximations for scaling up LVGPs and fully Bayesian inference for the LVGP hyperparameters.  We conduct numerical studies comparing plug-in inference against fully Bayesian inference over a few engineering models and material design applications. In contrast to previous studies on standard GP modeling that have largely concluded that a fully Bayesian treatment offers limited improvements, our results show that for LVGP modeling it offers significant improvements in prediction accuracy and uncertainty quantification over the plug-in approach.
\end{abstract}

\noindent%

\input{introduction}

\input{background}

\input{bayesian}

\input{empirical}

\input{conclusions}

\bibliography{references}

\input{supp}

\end{document}

%% file: introduction.tex
\section{Introduction}

Many scientific and engineering applications often require the use of surrogate models or emulators in various tasks such as optimization, sensitivity analysis, active learning, etc. Gaussian processes (GPs) are a popular class of surrogate models. Their principled uncertainty quantification (UQ) makes them useful for Bayesian optimization (BO) \cite{shahriari2016taking} and model calibration \cite{arendt2012quantification}. Traditionally, GP models have been developed for quantitative/numerical inputs. However, many applications involve one or more qualitative/categorical inputs. For example, in several material design applications, a goal is to find material compositions (e.g. atomic compositions) that have the desired target properties (such as resistivity, bandgap, etc.). 

Previous work that has developed GP models for systems involving one ore more qualitative input(s) include \cite{qian2008gaussian,han2009prediction,zhou2011simple,deng2017additive,zhang2019lvgp}, and \cite{roustant2020group}. Among them, the latent variable Gaussian process \cite{zhang2019lvgp} (LVGP, not to be confused with Gaussian process latent variable models \cite{lawrence2003gaussian,titsias2010bayesian} that are used for performing nonlinear dimensionality reduction) has typically achieved comparatively better modeling performance for such systems. When used within a BO framework, it has been applied to several material design and engineering design problems \cite{iyer2019data,zhang2020bayesian,iyer2020data,wang2020featureless,pelamatti2021bayesian} and has yielded improved results. The LVGP method maps the levels of each qualitative input to a set of numerical values for some latent numerical variables. The latent variable values for each qualitative factor quantify the ``distances” between the different levels, and therefore can be treated the same as numerical inputs in a GP model. In addition to the improved modeling performance, the latent variable mapping of the qualitative factors provides an inherent ordering and structure for the levels of the factor(s), which can provide insights into the effects of the qualitative factors on the response (see the examples in \cite{zhang2019lvgp}).

The latent variables that represent the levels are treated as unknown and must be estimated along with other GP hyperparameters before being used for prediction or UQ. By ``UQ", we essentially mean a prediction interval on the predicted response, as a function of the inputs. A common strategy for inference in standard GPs is to plug in point estimates, such as maximum likelihood estimates or the maximum a-posteriori (MAP) estimates, into the different quantities of interest such as the expected improvement sampling criterion for BO.  This plug-in approach, however, does not account for the uncertainty in the estimation of these hyperparameters. On the other hand,  fully Bayesian inference, where one marginalizes over the posterior distribution of the hyperparameters, takes this uncertainty into account in a principled manner (for e.g., \cite{handcock1993bayesian}). However, as it is more computationally expensive than plug-in inference, the plug-in approach is more commonly used.  For standard GPs with numerical inputs, the benefits of the fully Bayesian approach appear to be mixed. Some works (for e.g., \cite{helbert2009assessment,pmlr-v118-lalchand20a}) have found the fully Bayesian approach to significantly improve performance, while some others (for e.g., \cite{MINASNY2011150,chen2016analysis,chen2017flexible}) have found the uncertainty in the estimated parameters to contribute relatively little to the total uncertainty in the predicted response. However, fully Bayesian inference has been found to be more robust for BO applications \cite{forrester2008global,benassi2011robust}, where small initial designs are commonly used.

Unlike for the standard GP model, the effect of the estimating the latent variables from data on the performance of the LVGP model is yet to be studied.  Prior works on LVGP modeling, including ones for BO applications, have all used a plug-in approach with maximum likelihood estimates. The estimation uncertainty can be especially significant for the latent variables, whose numbers are usually much larger than that of the GP hyperparameters. This is of particular relevance for small initial datasets that are often encountered in material design applications (see \cite{balachandran2016adaptive}, e.g.), especially when one or more qualitative variables have many levels, in which case there will typically be some levels for which no response observations are available.  Improving the quality of UQ improves the performance of BO algorithms in these applications, which is a primary motivation for this work. 

In this work, we develop a fully-Bayesian approach for LVGP modeling and conduct numerical studies comparing it to plug-in inference on a few of material design applications that motivated this work, and also on a few engineering models. This involves developing appropriate prior distributions for the latent variables and the other hyperparameters. The fully Bayesian approach complicates the interpretation of the latent variables, since there is a different latent variable mapping for each Markov chain Monte Carlo (MCMC) draw. Consequently, we develop an approach for finding a single common latent variable mapping in order to capitalize on the interpretability advantages of LVGP modeling. Finally, we also develop approximations for scaling fully Bayesian inference for LVGPs to much larger data sets than is possible with exact inference.

One important finding is that a fully Bayesian treatment appears to be substantially more beneficial in LVGP modeling with qualitative inputs than in standard GP modeling with only numerical inputs. Whereas the improvements in standard GP modeling have been mixed (see above discussion), we observed substantial improvements for the fully Bayesian approach in every example that we considered. The improvement in the UQ (as measured by the mean interval score metric \cite{gneiting2007strictly}) was substantial in every example we considered. Moreover, and somewhat surprisingly, the predictive accuracy (as measured by the mean square error in predicting test response cases) was often improved substantially and never substantially worsened.  We, therefore, advocate for a fully Bayesian treatment of the LVGP hyperparameters, especially when one or more qualitative inputs have many levels. 

The remainder of the paper is structured as follows. We review GPs with numerical inputs and LVGPs in Section \ref{sec:methods}. We derive the fully Bayesian inference for the LVGP model in Section \ref{sec:bayesian}. We also discuss the ambiguity in interpreting the latent spaces under fully Bayesian inference, and provide a solution for the same.  In Section \ref{sec:sparse-lvgp}, we develop approximations for scaling up LVGPs, and fully Bayesian inference for the LVGP hyperparameters using sparse inducing point methods. We discuss the numerical studies in Section \ref{sec:empirical}. Finally, we offer concluding remarks in Section \ref{sec:conc}.

%% file: background.tex
\section{Background and notation} \label{sec:methods}
In this section, we review the standard GP model for quantitative inputs, related estimation and inference methods, and the LVGP model to handle qualitative factors. Let $y(\cdot)$ denote the true physical response surface model with inputs $\mathbf{w} = \smb{\mathbf{x},\mathbf{t}}$, where $\mathbf{x} = \smb{x_1,\ldots,x_I} \in \mathcal{X} \subset \mathbb{R}^I$ represents $I$ quantitative variables and $\mathbf{t} = \smb{t_1,\ldots,t_J}$ represents $J$ qualitative variables with $L_1,\ldots,L_J$ levels respectively. Without loss of generality, we assume that the $L_j$ levels of the $j^\mathrm{th}$ qualitative factor $t_j$ are coded as a $\{1,2,\ldots,L_j\}$. Let $\mathcal{W} = \mathcal{X} \times \prod_{j=1}^{J}\{1,\ldots,L_j\}$ denote the product input space.

\subsection{GP regression}

We consider GP regression models of the form:
\begin{equation}
    y\smb{\mathbf{w}} = f\smb{\mathbf{w}} + \epsilon,
\end{equation}

\noindent $f\smb{\cdot}$ is a GP, and $\epsilon$ is the zero-mean Gaussian noise term with variance $\sigma^2_\epsilon$. GPs define a prior over a space of functions $f:\mathcal{W}\xrightarrow{}\mathbb{R}$ with the property that any finite number of function values $f(\mathbf{w}_1),\ldots,f(\mathbf{w}_n)$ associated with different locations $\mathbf{w}_1,\ldots,\mathbf{w}_n \in\mathcal{W}$, have a joint Gaussian distribution. GPs are completely characterized by a mean function $\mu:\mathcal{W}\xrightarrow{}\mathbb{R}$ and a covariance or kernel function $k:\mathcal{W}\times\mathcal{W}\xrightarrow{}\mathbb{R}$. A common choice for the mean function is a constant $\mu$, which we will assume throughout this paper. The kernel function encodes assumptions about the function's properties and the choice is problem dependent.  A common choice for numerical inputs is the squared-exponential kernel
\begin{equation}
    k\smb{\mathbf{x},\mathbf{x}'} = \sigma^2\exp\smb{-\frac{1}{2}\sum_{i=1}^{I} \frac{\smb{x_i -x'_i}^2}{\omega_i^2}},
    \label{eq:se_kernel}
\end{equation}

\noindent where $\sigma^2$ is the prior variance hyperparameter, and $\omega_1,\ldots,\omega_I$ are positive length-scale hyperparameters.

Suppose we have a set of $N$ observations $\mathbf{Y} = \smb{y_1,\ldots,y_N}$ (e.g., from N runs of some complex code that simulates a physical system) at inputs $\textbf{W} = \left[\textbf{w}_1,\ldots,\textbf{w}_N\right]^\mathsf{T}$. By the marginalization properties of jointly Gaussian variables, the posterior distribution of $f\smb{\textbf{w}_*}$ at input $\textbf{w}_*$ conditioned on these observations, is again Gaussian with mean and variance given by:
\begin{align}
\widehat{f}\smb{\mathbf{w}_*;\btheta} = \Exp{f(\textbf{w}_*)|\textbf{W},\textbf{Y},\btheta} &= \mu_{\btheta}\mathbf{1}_N + \mathbf{k}\smb{\textbf{W},\mathbf{w}_*;\btheta}^T\bSigma\smb{\btheta}^{-1}\smb{\textbf{Y}- \mu_{\btheta}\mathbf{1}_N} \label{eq:gpmean} \\
\widehat{\sigma^2}\smb{\mathbf{w}_*;\btheta} = \Var{f(\mathbf{w}_*)|\mathbf{W},\textbf{Y},\btheta} &= \sigma^2_{\btheta}- \mathbf{k}\smb{\mathbf{W},\mathbf{w}_*;\btheta}^T\bSigma\smb{\btheta}^{-1}\mathbf{k}\smb{\mathbf{W},\mathbf{w}_*;\btheta}, \label{eq:gpvar}
\end{align}

\noindent where $\btheta$ is the vector of hyperparameters including the noise variance $\sigma^2_{\epsilon,\btheta}$, $\bSigma\smb{\theta}$ is the covariance matrix of the observations whose element at the $n^\mathrm{th}$ row and $m^\mathrm{th}$ column is $k\smb{\mathbf{w}_n,\mathbf{w}_m;\btheta} + \sigma^2_{\epsilon,\btheta}\delta_{nm}$, $\mathbf{k}\smb{\mathbf{W},\mathbf{w}_*;\btheta}=  \left[k\smb{\mathbf{w}_1,\mathbf{w}_*;\btheta}, \ldots, k\smb{\mathbf{w}_N,\mathbf{w}_*;\btheta}\right]^T$ is the $N$ dimensional covariance vector between $f\smb{\mathbf{w}_*}$ and the observations, and $\mathbf{1}_N$ is a $N\times 1$ vector of ones.

The GP hyperparameters $\btheta$ are usually unknown in practice. A common strategy is to use a \textit{plug-in Bayes} approach, also known as \textit{Empirical Bayes}, where an estimate $\widehat{\btheta}_N$ is first computed from data, and then plugged into the posterior predictive distribution. Under this plug-in approach, the posterior distribution of $f\smb{\mathbf{w}_*}$ is again Gaussian whose mean and variance are calculated by plugging in $\widehat{\btheta}_N$ into \eqref{eq:gpmean} and \eqref{eq:gpvar} respectively. A common strategy for estimation is to optimize $\btheta$ with respect to either the log-likelihood
\begin{equation}
    \mathcal{L}\smb{\btheta} = -\frac{N}{2} \ln 2\pi - \frac{1}{2} \ln |\bSigma\smb{\btheta}| - \frac{1}{2} \smb{\textbf{Y}- \mu\smb{\mathbf{W}}}^T \bSigma\smb{\btheta}^{-1}
    \smb{\textbf{Y}- \mu\smb{\mathbf{W}}},
    \label{eq:likelihood}
\end{equation}

\noindent which gives the maximum likelihood estimate, or the log-posterior,
\begin{equation}
    \log p\smb{\btheta|\mathbf{W},\mathbf{Y}} = \mathcal{L}\smb{\btheta} + \log p\smb{\btheta} + \mathrm{const.},
\end{equation}
which gives the \textit{maximum a-posteriori} (MAP) estimate. Here, $p\smb{\btheta}$ is the prior distribution density over $\btheta$.

A more rigorous approach to obtain the posterior predictive distributions is to use \textit{fully Bayesian} inference, where one marginalizes over the posterior distribution of $\btheta$ given data. The density of the posterior distribution of $f\smb{\mathbf{w}_*}$ given the data is 
\begin{equation}
    p\smb{f\smb{\mathbf{w}_*}|\mathbf{W},\mathbf{Y}} = \int
    p\smb{f\smb{\mathbf{w}_*}|\mathbf{W},\mathbf{Y},\btheta}
    p\smb{\btheta|\mathbf{W},\mathbf{Y}}d\btheta.
    \label{eq:post-bayes-dist}
\end{equation}

There is usually no closed form expression for the above distribution, as the integrals involved are often intractable. In practice, one often uses MCMC methods to draw a finite number of samples from the distribution $p\smb{\btheta|\mathbf{W},\mathbf{Y}}$, which can then be used to compute different quantities of interest. In this work, we use the No-U-Turn-Sampler \cite{hoffman2014no} algorithm for drawing the MCMC samples.

\subsection{LVGP models for systems with qualitative inputs}

Standard GP correlation functions, such as the squared exponential, Mat\'ern, power exponential, etc. cannot be directly applied to qualitative inputs because there is no inherent distance metric in the qualitative space. The recently proposed LVGP approach \cite{zhang2019lvgp} provides a natural and convenient way to handle qualitative inputs  by mapping the levels of each qualitative factor onto low-dimensional continuous latent spaces. To be more specific, consider a qualitative input $t$ with $L$ levels. The LVGP approach constructs a mapping $\mathbf{z}:\{1,\ldots,L\} \xrightarrow{} \mathbb{R}^d$, where $d < L$ is the dimension of the latent variable (LV) space. With this mapping, the LVGP approach defines kernels of the form
\begin{equation}
    k\smb{t,t'} = \widetilde{k}\smb{\mathbf{z}\smb{t},\mathbf{z}\smb{t'}},
\end{equation}

\noindent where $\widetilde{k}$ is any kernel function on $\mathbb{R}^d$. 

\begin{figure}
    \centering
    \includegraphics[width=0.7\linewidth]{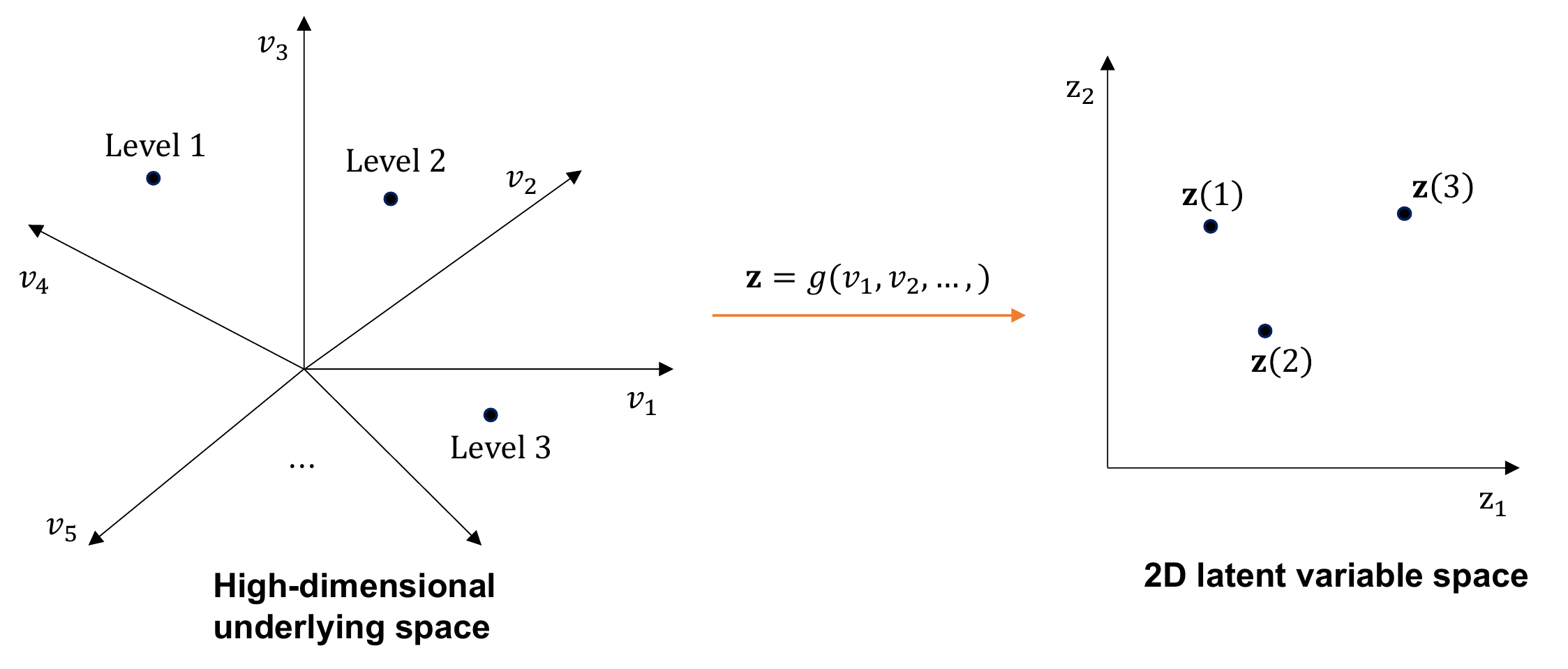}
    \caption{2D latent variable mapping for a single qualitative factor $t$ with three levels}
    \label{fig:lvgp_illustration}
\end{figure}

The LVGP is based on the premise that for any real physical system, the effects of a qualitative input $t$ must be due to some underlying (and perhaps very high-dimensional) quantitative variables $\{v_1,v_2,\ldots\} = \{v_1\smb{t},v_2\smb{t},\ldots\}$. Consider the situation in Figure \ref{fig:lvgp_illustration}, where the three levels of $t$ are associated with points on the high-dimensional space of $\{v_1,v_2,\ldots\}$. The LVGP uses sufficient dimensionality reduction arguments to \textit{implicitly} construct a low dimensional representation $\mathbf{z}\smb{t} = g\smb{v_1\smb{t},v_2\smb{t},\ldots}$ that accounts for most of their effects. Based on empirical studies, \cite{zhang2019lvgp} argued that $d=2$ was often a good choice.
%, and we adopt this convention also, although the approach is easily extended to $d > 2$. 

When there are multiple qualitative inputs, a separate latent space is used for each qualitative input. Let $\mathbf{z}^{(j)}: \{1,\ldots,L_j\} \xrightarrow{} \mathbb{R}^{d_j}$ denote the $d_j$-dimensional LV mapping for the $j^\mathrm{th}$ qualitative input $t_j$. If the squared exponential kernel is used for both the quantitative variables and the LVs (i.e., $\widetilde{k}\smb{\mathbf{z}\smb{t},\mathbf{z}\smb{t'}} = \exp\left[-\frac{1}{2} \left\lVert\mathbf{z}\smb{t} - \mathbf{z}\smb{t'}\right\rVert^2 \right]$) for the qualitative inputs, the covariance kernel for the LVGP is given by
\begin{equation}
    k\smb{\mathbf{w},\mathbf{w}'} = \sigma^2 \exp\left[
    -\frac{1}{2}\sum_{i=1}^{I} \frac{\smb{x_i -x'_i}^2}{\omega_i^2} 
    -\frac{1}{2}\sum_{j=1}^{J} \left\lVert\mathbf{z}^{(j)}\smb{t_j} - \mathbf{z}^{(j)}\smb{t'_j}\right\rVert^2\right].
    \label{eq:lvgp-secovariance}
\end{equation}

\noindent In the above, the length scales for the LVs are set to unity. This is because these length scales are implicitly estimated along with the LV mapping. 

The LVs are additional hyperparameters that must be estimated along with the other GP hyperparameters. Existing works in the literature employing the LVGP have all used maximum likelihood estimation followed by the plug-in Bayes approach for generating predictions. The impact of estimation of LVs on the performance of the LVGP model is yet to be studied. In the remainder of the paper, we demonstrate that the impact of the estimation uncertainty in the LVs can be substantial, and that the fully Bayesian approach that we develop for LVGPs appropriately accounts for the uncertainty, improving both the prediction accuracy and the UQ.

%% file: bayesian.tex
\section{Fully Bayesian inference for LVGPs} \label{sec:bayesian}

% The plug-in Bayes approach can be viewed as approximating  $p\smb{\btheta|\mathbf{W},\mathbf{Y_N}}$ by a Dirac measure at $\btheta_N$. This is justified only if the distribution is \textit{sufficiently} concentrated around $\btheta_N$. Moreover, plug-in Bayes will ignore the uncertainty in estimating the hyperparameters, which can be significant. This is relevant in BO applications, which typically use small datasets, and the data can sometimes result in misleading estimates which leads to poor performance \cite{benassi2011robust}. In contrast, fully Bayesian inference offers a principled method to quantify this uncertainty. However, running MCMC to convergence can be much more computationally expensive than obtaining a point estimate.  

In this section, we discuss various aspects for fully Bayesian inference for LVGPs -  the choice of the prior distributions (Section \ref{sec:priors}), our choice of the MCMC algorithm and obtaining predictions from the model (Section \ref{sec:predictions}), and interpreting the estimated LVs (Section \ref{sec:interpret}). As in \cite{zhang2019lvgp}, we will assume that the kernel over the LVs is stationary. 

\subsection{Priors for the latent variables} \label{sec:priors}
For fully Bayesian inference, we need to specify prior distributions for the LVs and for the other GP hyperparameters. In this section, we will discuss our choice of the prior distribution for the LVs. Prior distributions used for the other GP hyperparameters are standard choices. In the following, we drop the superscript $(j)$ from the LVs for notational convenience. For a qualitative variable $t$ with $L$ levels and its LV mapping $\mathbf{z}$, let $\mathbf{z}\smb{l} = \smb{z_{l1},\ldots,z_{ld}}$ denote the mapped values for level $l \in \{1,\ldots,L\}$.

The prior on the LVs for a qualitative variable $t$ can be used to encode any domain knowledge (if available) about the similarity between the different levels of $t$ in terms of their effects on the response. This domain knowledge may be available in multiple forms. For example, we could be given an incomplete set of numerical descriptors that are either known to or suspected to have significant effects on the response. The Euclidean distances between these descriptor values could potentially be used to be define some appropriate prior similarity measure for the LVs. However, in this work we focus on the more general setting where no such domain knowledge exists.

In the absence of any domain information, we define the following \textit{weakly informative} prior over the LVs for $t$:
% for the $r^\mathrm{th}$ latent variable for the $l^\mathrm{th}$ level:
\begin{align}
    z_{lr} \overset{\mathrm{i.i.d.}}{\sim} \mathcal{N}\smb{0,\frac{1}{L\gamma}} \quad \forall l \in \{1,\ldots,L\}, r \in \{1,\ldots,d\},
    \label{eq:lv-prior}
\end{align}

\noindent where $\gamma$ is a precision hyperparameter that is common for all LVs for $t$, and is jointly inferred along with these LVs. Note that different qualitative factors have different precision hyperparameters.  The prior \eqref{eq:lv-prior} encourages the values of the LVs to be concentrated towards 0. This has an indirect effect of shrinking the distances between the different levels in the LV space, and can be viewed as a form of regularization of the effects of  $t$. The degree of regularization for $t$ is controlled by the precision $L\gamma$. The factor of $L$ is included to account for the fact that the number of LVs to be estimated increases with $L$ and a larger regularization will be needed for larger values of $L$. 

Alternatively, one can place a completely non-informative prior on the LVs. In practice, one can approximate this via i.i.d. uniform priors with relatively large lower and upper limits (e.g., -10 and 10, respectively). With a completely non-informative prior, there is effectively no regularization on the effects of the qualitative variable. We find that the completely non-informative priors negatively impacted MCMC convergence, irrespective of the choice of the MCMC algorithm. We will, therefore, restrict our attention to the weakly informative prior \eqref{eq:lv-prior}.

% add the rotational invariance problems

% \cite{zhang2019lvgp} introduce additional constraints on the LVs to remove translation and rotation invariances. In particular, the  

\begin{figure}[tbhp]
    \centering
    \includegraphics[width=0.45\textwidth]{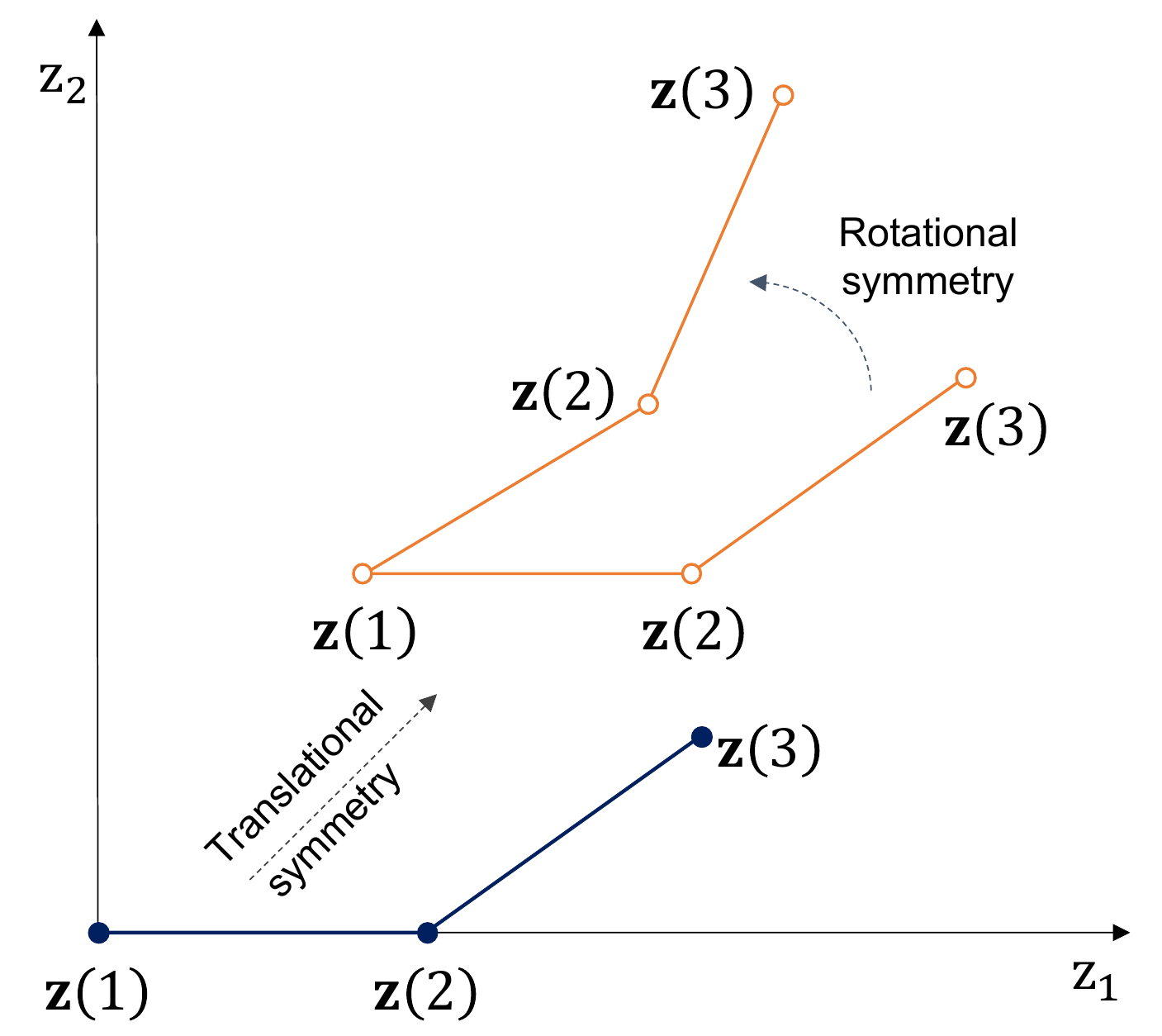}
    \caption{Translation and rotation symmetries in a 2D LV space under the LVGP likelihood: Three different sets of mapped LVs ($\mathbf{z}\smb{1},\mathbf{z}\smb{2},\mathbf{z}\smb{3}$) have the same pairwise distances, and hence result in the same pairwise covariances.}
    \label{fig:lvsymmetries}
\end{figure}

With a stationary kernel over the LVs,  the correlations between the different levels is dependent only on the pairwise distances between the corresponding LV values of the levels. Therefore, unconstrained LVs are subject to translation and rotation symmetries under the LVGP likelihood (see Figure \ref{fig:lvsymmetries} for illustration in a 2D LV space). While the prior \eqref{eq:lv-prior} reduces translation symmetries in the LVs under the posterior, rotation symmetries still exist - the two sets of LVs $\{\mathbf{z}\smb{l}:l\in\{1,\ldots,L\}\}$ and
$\{\mathbf{Q}^\mathsf{T}\mathbf{z}\smb{l}:l\in\{1,\ldots,L\}\}$ have the same posterior probabilities, where $\mathbf{Q}$ is any orthogonal matrix in $\mathbb{R}^d$. This rotational symmetry in the LVs can potentially reduce sampling efficiencies of the MCMC scheme, which would necessitate large number of MCMC draws, thereby increasing the computational expense of fully Bayesian inference.
To eliminate these symmetries, \cite{zhang2019lvgp} fix the coordinate frame of reference. For example, in a 2D LV space, they constrain the mapped LVs for the first level, $\mathbf{z}\smb{1}$, at the origin, and those of the second level, $\mathbf{z}\smb{2}$, to lie on the horizontal axis. 
%This also reduces the number of free parameters from $2L$ to $2L-3$.
For a general $d$,  they set $\mathbf{z}\smb{1} = \mathbf{0}_{d}$,  $\mathbf{z}\smb{2} = \begin{bmatrix}
    z_{21} & \mathbf{0}_{d-1}
\end{bmatrix}$, and $\mathbf{z}\smb{l} = \begin{bmatrix}
    z_{l1} & \ldots & z_{l\smb{l-1}} & \mathbf{0}_{d-l+1}
\end{bmatrix}$ for $2 < l\leq d$. This results in $d\smb{d-1}/2$ fewer free parameters.
However, a challenge with this parameterization for fully Bayesian inference is that the prior distribution cannot represent independent and identically distributed LVs, because this parameterization treats the first $d$ levels differently from the other levels. For example, under independent normal or uniform prior distributions with mean at 0, the LVs for any level would be closer on average to the first level than to any other level. This treatment of the LVs is problematic, because the ordering of the levels is arbitrary, and it can affect the quality of the LVGP model, especially with a small number of training observations, or when some of the levels are unobserved in the training data.

To rectify this, we instead modify the prior \eqref{eq:lv-prior} by first defining ``raw'' LV parameters $\{\widetilde{\mathbf{z}}\smb{l}=(\widetilde{z}_{l1},\ldots,\widetilde{z}_{ld}): l\in\{1,\ldots,L\}\}$, which are i.i.d. for all levels, and then transforming them to the coordinate frame of reference used by \cite{zhang2019lvgp} to obtain the actual LVs. We first consider the case of a 2D LV space. The resulting two-stage prior is as follows:
\begin{align}
    \widetilde{z}_{lr} \overset{\mathrm{i.i.d.}}{\sim} \mathcal{N}\smb{0,\frac{1}{L\gamma}} \quad \forall l \in \{1,\ldots,L\}, r \in \{1,2\},\\
    \mathbf{z}\smb{l} = 
    \mathbf{R}\smb{\phi\smb{\widetilde{\mathbf{z}}\smb{1},\widetilde{\mathbf{z}}\smb{2}}}^\mathsf{T}
    \smb{\widetilde{\mathbf{z}}\smb{l}-\widetilde{\mathbf{z}}\smb{1}} \quad \forall l \in \{1,\ldots,L\}, \label{eq:raw-to-lvs}
\end{align}

\noindent where 
\begin{equation}
    \phi\smb{\widetilde{\mathbf{z}}\smb{1},\widetilde{\mathbf{z}}\smb{2}} = \mathrm{tan}^{-1}\smb{\frac{\widetilde{z}_{22}-\widetilde{z}_{12}}{\widetilde{z}_{21}-\widetilde{z}_{11}}},
\end{equation}

\noindent and $\mathbf{R}\smb{\phi}$ is a rotation matrix in $\mathbb{R}^2$ parameterized by $\phi$ and given by
\begin{equation}
    \mathbf{R}\smb{\phi} = \begin{bmatrix}
    \cos \phi & -\sin\phi\\
    \sin \phi & \cos \phi
    \end{bmatrix}.
\end{equation}

\noindent In \eqref{eq:raw-to-lvs}, the raw LVs are first translated so that the first level is at the origin, and then rotated by $\mathbf{R}\smb{\phi\smb{\widetilde{\mathbf{z}}\smb{1},\widetilde{\mathbf{z}}\smb{2}}}$ so that the mapped values of the second level lie on the horizontal axis. Note that the mapped LVs for the remaining levels are not constrained to lie on either axis. The prior distribution on the actual LVs are, clearly, not i.i.d., although by symmetry arguments the distances between each pair of levels follow the same prior distribution. The modifications represented by \eqref{eq:raw-to-lvs} ensure that the results are theoretically invariant (barring numerical errors) to which two levels are chosen to lie at the origin and on the horizontal axis.

For $d>2$, the $\mathbf{R}\smb{\cdot}$ matrix in \eqref{eq:raw-to-lvs} is replaced by a product of $d(d-1)/2$ Givens rotation matrices in $d$ dimensions. Givens rotation matrices in $d$-dimensions, $\mathbf{R}_{ij}\smb{\phi_{ij}}$, are $d\times d$ matrices that take the form of the identity matrix except for the $(i,i)$ and $(j,j)$ positions which are replaced by $\cos\smb{\phi_{ij}}$, and the $(i,j)$ and $(j,i)$ positions which are replaced by $-\sin\smb{\phi_{ij}}$ and $\sin\smb{\phi_{ij}}$ respectively. The LVs for a general $d$ are given by
\begin{equation}
    \mathbf{z}\smb{l} = 
    % \smb{\prod_{i=1}^{d-1}\prod_{j=i+1}^{d}  \mathbf{R}_{ij}\smb{\phi_{ij}\smb{\widetilde{\mathbf{z}}\smb{1},\ldots,\widetilde{\mathbf{z}}\smb{j}}}^\mathsf{T}}
    \mathbf{R}^{(d)}\smb{\widetilde{\mathbf{z}}\smb{1},\ldots,\widetilde{\mathbf{z}}\smb{d}}^\mathsf{T}
    \smb{\widetilde{\mathbf{z}}\smb{l}-\widetilde{\mathbf{z}}\smb{1}} \quad \forall l \in \{1,\ldots,L\},
    \label{eq:raw-to-lvs-gen-d}
\end{equation}

\noindent where 
\begin{equation}
\mathbf{R}^{(d)}\smb{\widetilde{\mathbf{z}}\smb{1},\ldots,\widetilde{\mathbf{z}}\smb{d}} = \prod_{i=1}^{d-1}\prod_{j=i+1}^{d}  \mathbf{R}_{ij}\smb{\phi_{ij}\smb{\widetilde{\mathbf{z}}\smb{1},\ldots,\widetilde{\mathbf{z}}\smb{i+1}}},
\end{equation}

\noindent and $\phi_{ij}\smb{\widetilde{\mathbf{z}}\smb{1},\ldots,\widetilde{\mathbf{z}}\smb{i+1}}$ is set such that the $j^\mathrm{th}$ LV for the $\smb{i+1}^\mathrm{th}$ level to be zero. In practice, the matrix multiplication operations in  \eqref{eq:raw-to-lvs-gen-d} are implemented sequentially using Algorithm \ref{alg:lv-general-d}.

\begin{algorithm}
\caption{Computing the LVs from their corresponding raw parameters for $d>2$}\label{alg:lv-general-d}
\begin{algorithmic}[1]
 \STATE{Set $\mathbf{Z} := \begin{bmatrix}
    \mathbf{0}_d & \smb{\widetilde{\mathbf{z}}\smb{2}-\widetilde{\mathbf{z}}\smb{1}} & \ldots & \smb{\widetilde{\mathbf{z}}\smb{L}- \widetilde{\mathbf{z}}\smb{1}}
\end{bmatrix}^\mathsf{T}$}
\FORALL{$i \in \{1,\ldots,d-1\}$}
    \FORALL{$j \in \{i+1,\ldots,d\}$}
        \STATE{$\phi_{ij} = \tan^{-1}\smb{{Z_{\smb{i+1},j}}/{Z_{\smb{i+1},i}}}$}
        \STATE{Update $\mathbf{Z} :=  \mathbf{Z}\cdot\mathbf{R}_{ij}\smb{\phi_{ij}}$}
    \ENDFOR
\ENDFOR
\FORALL{$l \in \{1,\ldots,L\}$}
    \STATE{Set $\mathbf{z}\smb{l} = \begin{bmatrix}
        Z_{l,1} & Z_{l,2} & \ldots & Z_{l,d}
    \end{bmatrix}^\mathsf{T}$} 
\ENDFOR
\end{algorithmic}
\end{algorithm}

With this formulation, for a qualitative variable with $L$ levels and using a $d$-dimensional LV spave, there are $Ld+1$ parameters to estimate. And so, with $J$ qualitative variables, we have  $J(Ld+1)$ parameters to estimate.  
Algorithm \ref{alg:lv-general-d} has an $O(Ld^2)$ computational cost for a qualitative variable with L levels and d number of latent variable dimensions. However, the main computational cost in each posterior evaluation comes from computing the Cholesky factorization, which is $O(N^3)$.

Finally, for a prior distribution on $\gamma$, we consider the Gamma distribution. {The support of the Gamma distribution is the set of positive real numbers. Larger values of $\gamma$ result in larger amounts of regularization on the LVs.} The Gamma distribution has two additional hyperparameters, the concentration $\alpha$ and the rate $\beta$. 
{ 
We restrict $\alpha >1$ and set $\beta=\alpha-1$, so that the mode of the distribution is at $\gamma=1$.  This results in only a single hyperprior parameter. We find that the convergence and performance of the model is not sensitive to the choice of $\alpha$. Refer to Appendix \ref{sec:app-hyperprior} for more details. In our numerical experiments in Section \ref{sec:empirical}, we use $\alpha=2$.}

\subsection{Predictions and UQ} \label{sec:predictions}
In this section, we briefly discuss our choice of the MCMC algorithm for drawing samples from the posterior distribution of the LVGP hyperparameters, and then we state the formulae and procedures for obtaining predictions and confidence intervals on these predictions. 

The joint posterior distribution for the LVs and the other GP hyperparameters is typically high dimensional. The MCMC algorithm should have a few desired properties in order to efficiently converge to this posterior distribution. Firstly, the algorithm needs to suppress random walk behavior for the Markov chain to mix well in high dimensions. Otherwise, the chain will take too long to converge to the posterior, and thereby require many evaluations for the posterior. Secondly, if the algorithm has additional tuning parameters, there must be a mechanism to tune them during the burn-in or warm-up phase. Among the various algorithms that we have tested, the No-U-Turn-Sampler \cite{hoffman2014no} algorithm\footnote{as implemented by the NumPyro \cite{phan2019composable,bingham2019pyro} library in Python} satisfies both properties, and is found to work well for our case.

Under MCMC sampling, the posterior predictive distribution of $f\smb{\mathbf{w}_*}$ at input $\mathbf{w}_*$ is approximated as a Gaussian mixture distribution
\begin{equation}
    f\smb{\mathbf{w}_*}|\mathbf{W},\mathbf{Y}
    \sim \frac{1}{B} \sum_{i=1}^{B} \mathcal{N}\smb{\widehat{f}\smb{\mathbf{w}_*;\btheta_b},\widehat{\sigma^2}\smb{\mathbf{w}_*;\btheta_b}},
    \label{eq:mcmc-post-dist}
\end{equation}

\noindent where $\left\{\btheta_1,\ldots,\btheta_B\right\}$ are MCMC samples drawn from $p\smb{\btheta|\mathbf{W},\mathbf{Y}}$, and $\widehat{f}\smb{\mathbf{w}_*;\btheta}$ and $\widehat{\sigma^2}\smb{\mathbf{w}_*;\btheta_b}$ are defined in \eqref{eq:gpmean} and \eqref{eq:gpvar} respectively. The mean and variance of this distribution are straightforward to compute and are as follows:
\begin{align}
\Exp{f\smb{\mathbf{w}_*}|\mathbf{W},\mathbf{Y}} &= 
\frac{1}{B}\sum_{b=1}^{B}\widehat{f}\smb{\mathbf{w}_*;\btheta_b} \\
\mathrm{Var}\left[f\smb{\mathbf{w}_*}|\mathbf{W},\mathbf{Y}\right] &= \frac{1}{B}\sum_{b=1}^{B}\widehat{\sigma^2}\smb{\mathbf{w}_*;\btheta_b} + \nonumber\\ & \frac{1}{B}\sum_{b=1}^{B}\smb{\widehat{f}\smb{\mathbf{w}_*;\btheta_b} - \Exp{f\smb{\mathbf{w}_*}|\mathbf{W},\mathbf{Y}}}^2. \label{eq:post-variance}
\end{align}

In this work, we define UQ in terms of prediction intervals on the predicted response (as a function of the inputs). Unfortunately, there are no analytical expressions for the quantiles of a Gaussian mixture distribution. So, the variance term \eqref{eq:post-variance} cannot be used to directly compute the quantiles. Instead, we estimate the quantiles empirically by drawing samples from this distribution as shown in Algorithm \ref{alg:conf}. When computing the confidence intervals in our experiments in Section \ref{sec:empirical}, we use about $M=10000$ samples. This is likely overkill, and we could obtain sufficiently accurate confidence intervals using as few as 500-1000 samples. However, the confidence interval computations are very fast, in part because the operations in Algorithm \ref{alg:conf} can be easily vectorized. So, we erred on the side of caution and used a large number of samples.

\begin{algorithm}
\caption{95\% prediction intervals for $f\smb{\mathbf{w}_*}|\mathbf{W},\mathbf{Y}$}\label{alg:conf}
\begin{algorithmic}[1]
\STATE Draw $M$ samples $f_1^*,\ldots,f_M^*$ from the distribution \eqref{eq:mcmc-post-dist}
\STATE Sort the samples in ascending order: $f_{(1)}^*\leq\ldots,f_{(M)}^*$
\STATE Estimate the 2.5$^\mathrm{th}$ percentile $f_{(q_l)}^*$, where $q_l = \lceil \frac{2.5}{100}\times M \rceil$
\STATE Estimate the 97.5$^\mathrm{th}$ percentile $f_{(q_u)}^*,$ where $q_u = \lceil \frac{97.5}{100}\times M \rceil$
\STATE Return $\left[f_{(q_l)}^*,f_{(q_u)}^*\right]$
\end{algorithmic}
\end{algorithm}

\subsection{Latent space interpretation} \label{sec:interpret}
\begin{figure}[htb]
    \centering
    \includegraphics[width=0.9\linewidth]{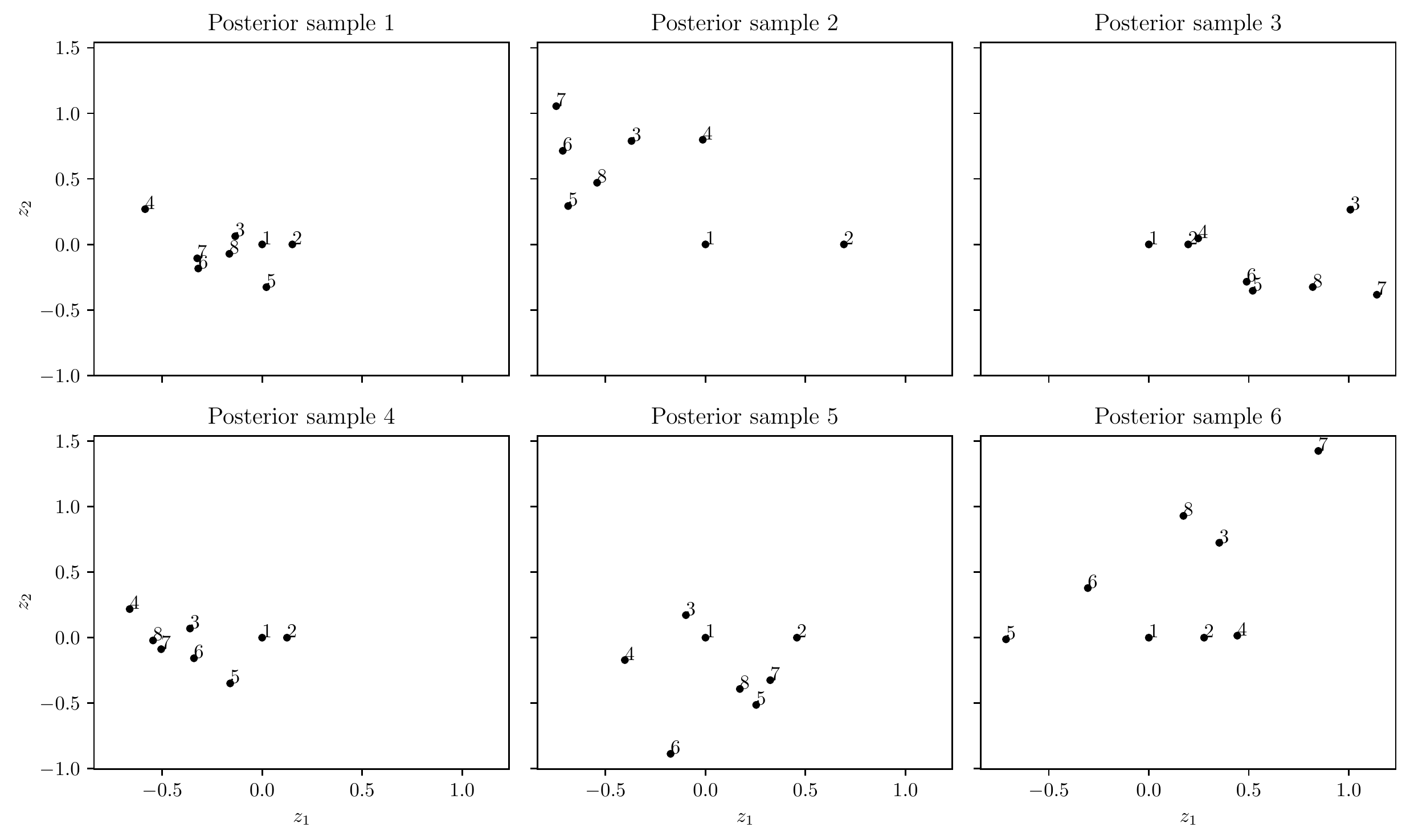}
    \caption{Examples of different estimated latent spaces corresponding to different MCMC samples from the posterior}
    \label{fig:post-latents}
\end{figure}

The LV mapping provides an inherent ordering and structure for the different levels of a qualitative input, in terms of their effects on the response. Therefore, a secondary benefit of the LVGP approach is to provide an interpretation of the effects of these different levels. Interpreting the LVs with fully Bayesian inference is, however, less straightforward. As shown in Figure \ref{fig:post-latents}, there are multiple latent spaces, each corresponding to a different MCMC sample drawn from the posterior. Since the estimation uncertainty in the LVs can be large, especially with small training data sets, the variation in the MCMC latent space samples can be large. {Due to this sample-to-sample variability, the different latent space samples can result in different interpretations of the relations among the levels,}
thereby making interpretation of the effects of the qualitative variable challenging. 

To resolve this ambiguity in interpretation and to help average out sample-to-sample variability, we propose to find a common latent space that reconciles the differences between the different latent spaces by minimizing a discrepancy measure over these different latent spaces. Since the latent variables are ultimately used to compute the covariance matrix over the levels, we define the discrepancy measure in terms of differences in their resulting covariance matrices. In the following, we drop the subscript $j$ from the LVs for notational convenience. For a LV mapping $\mathbf{z}$ for a qualitative input $t$ with $L$ levels, let $\mathbf{Z} = \left[\mathbf{z}\smb{1} \quad \mathbf{z}\smb{2} \quad \ldots \quad \mathbf{z}\smb{L} \right]^\mathsf{T}$ denote the corresponding $L\times d$ LV matrix. Let $\widetilde{\mathbf{k}}\smb{\mathbf{Z}}$ denote the $L\times L$ covariance matrix across levels corresponding to the LV matrix $\mathbf{Z}$, where the element at the row $l_1$ and column $l_2$ is $\widetilde{k}\smb{\mathbf{z}\smb{l_1}, \mathbf{z}\smb{l_2}}$. We define the discrepancy measure between two LV matrices $\mathbf{Z}$ and $\mathbf{Z}'$ (e.g., corresponding to two different MCMC draws of the LVs) as $\left\Vert\widetilde{\mathbf{k}}\smb{\mathbf{Z}}-\widetilde{\mathbf{k}}\smb{\mathbf{Z}'}\right\Vert_F$, where $\left\Vert\cdot\right\Vert_F$ is the Frobenius norm. 

Let $\mathbf{Z}_{(1)},\ldots,\mathbf{Z}_{(B)}$ be the $B$ latent variable matrices corresponding to the $B$ different hyperparameter samples from the posterior. We find a \textit{representative} latent variable matrix $\mathbf{Z}_* \in \mathbb{R}^{L\times d}$ as
\begin{align}
    \mathbf{Z}_* = \underset{\mathbf{Z}}{\arg\min} \quad  & \frac{1}{B}\sum_{b=1}^{B}\left\Vert\widetilde{\mathbf{k}}\smb{\mathbf{Z}_{(b)}}-\widetilde{\mathbf{k}}\smb{\mathbf{Z}}\right\Vert_F \label{eq:representative-lv}
    \\
    % \mathrm{s.t.} \quad & Z_{11} = 0, Z_{12}=0, Z_{21} = 0,
    \mathrm{s.t.} \quad & Z_{l,r} = 0 \quad \forall l \in \{1,\ldots,d\}, r \in \{l,\ldots,d\} \label{eq:rep-const1}
    %\\ \quad&  \mathbf{Z}_{lr} \geq 0 \quad \forall l \in \{2,\ldots,d\}, r \in \{l-1,\ldots,d-1\}.  \label{eq:rep-const2}
\end{align}

\noindent where $Z_{l,r}$ is the $r^\mathrm{th}$ LV for the $l^\mathrm{th}$ level. Constraints \eqref{eq:rep-const1} fix the coordinate frame of reference to the one used by \cite{zhang2019lvgp} to deal with the translation and rotation symmetries in the LVs, which were discussed in Section \ref{sec:priors}.

Once $\mathbf{Z}_*$ is obtained, the representative LV space can be interpreted in a similar way as the LV space obtained from point estimates. For systems with multiple qualitative inputs, the optimization problem \eqref{eq:representative-lv} is run separately for each qualitative input.

\begin{figure}[tbhp]
    \centering
    \subfloat[True numerical space]{\label{fig:borehole-true}\includegraphics[width=0.32\textwidth]{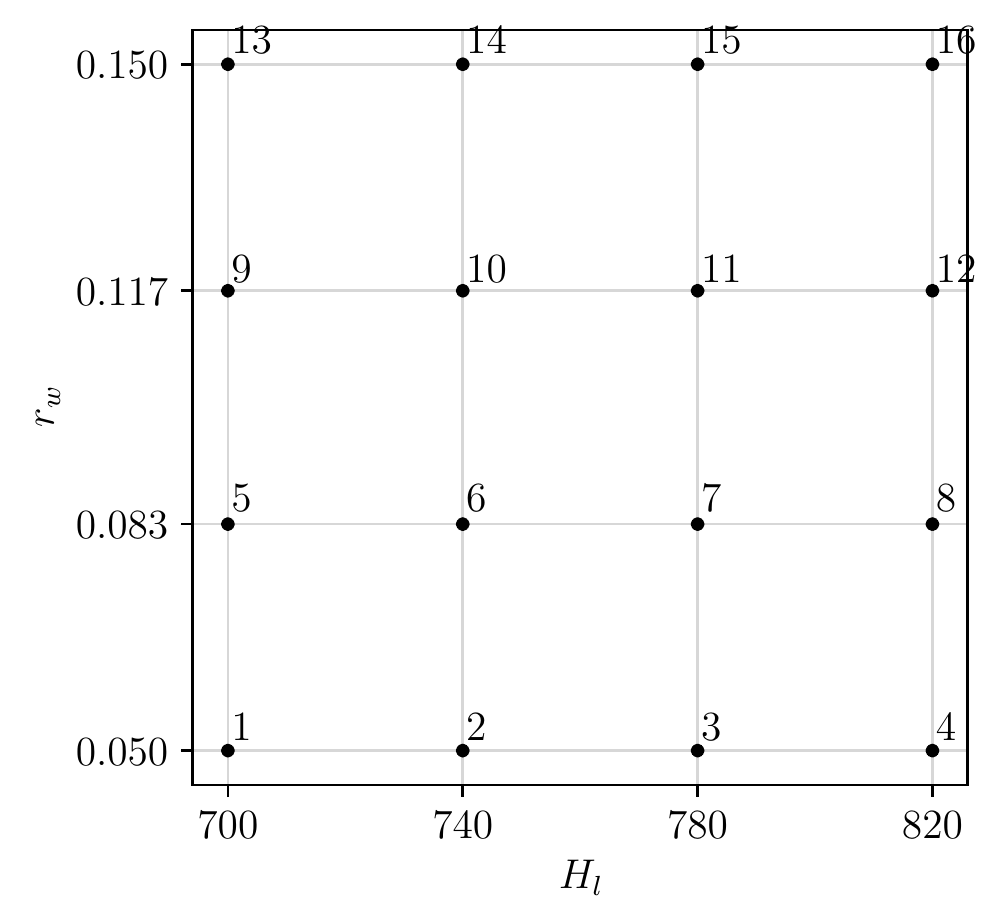}}
    \subfloat[32 training obs.]{ \label{fig:borehole-32}\includegraphics[width=0.32\textwidth]{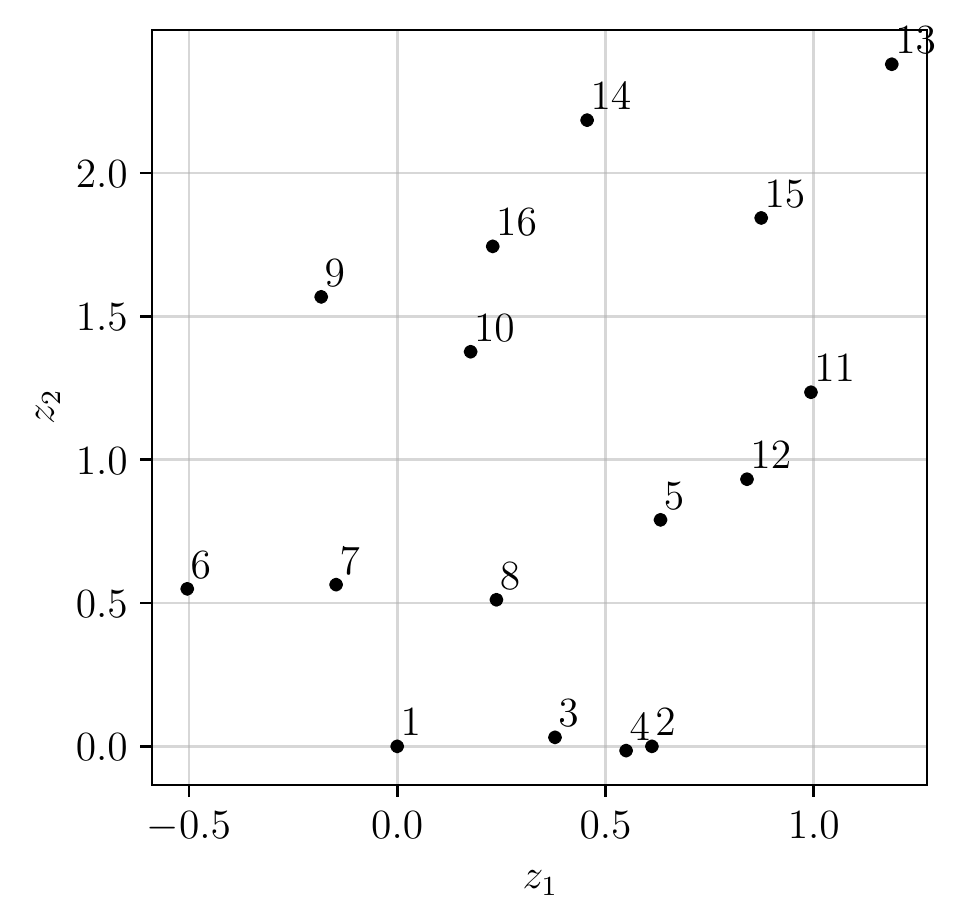}}
    \subfloat[64 training obs.]{ \label{fig:borehole-64}\includegraphics[width=0.32\textwidth]{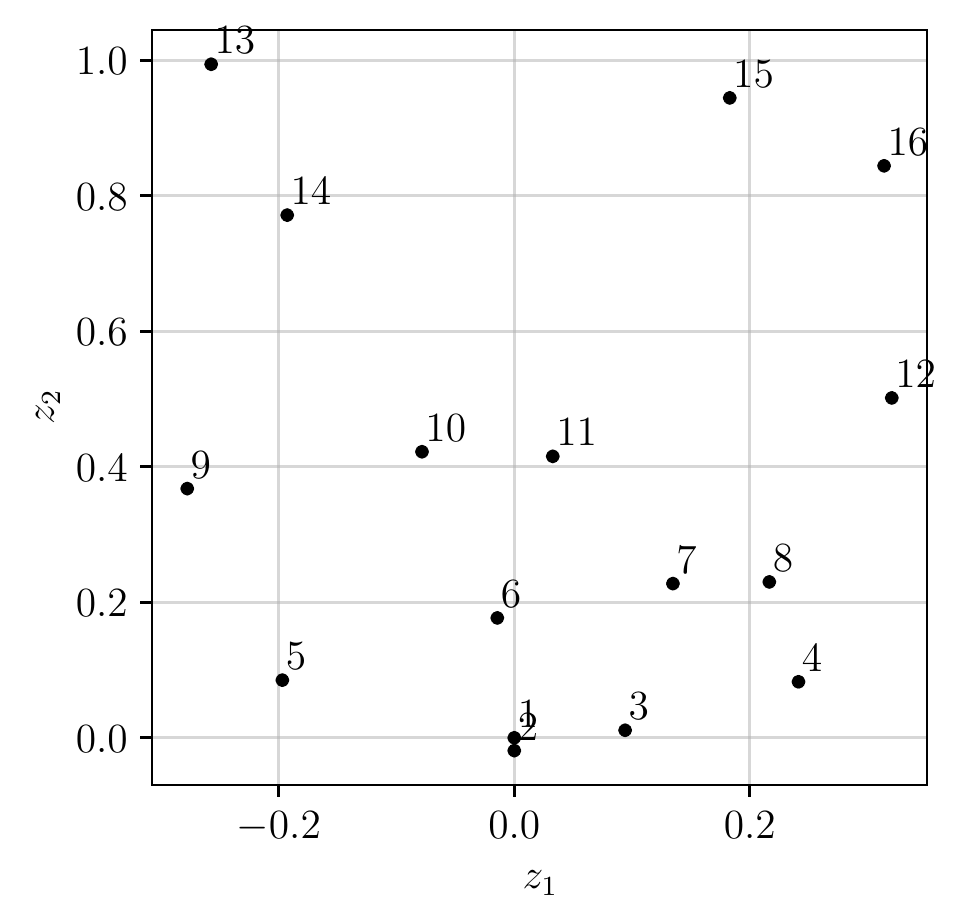}}
    \caption{For the borehole example, (a) the true numerical space (b) representative LV space with 32 observations, (c) representative LV space with 64 observations observations}
    \label{fig:borehole-32-latents}
\end{figure}

For illustration, consider the borehole function, that is commonly used to illustrate GP modeling \cite{borehole}. The flow rate of water through a borehole, that is drilled from the ground surface through two aquifers is 

\begin{equation*}
    2\pi T_u\smb{H_u-H_l}\smb{
    \log\smb{\frac{r}{r_w}}\smb{
        1+ 2\frac{LT_u}{\log\smb{r/r_w}r_w^2K_w} + \frac{T_u}{T_l}
    }
    }^{-1},
\end{equation*}

\noindent where the 8 inputs are ($T_u,r,r_w,H_u,T_l,H_l,L,K_w$). We modify this model by first discretizing two underlying numerical variables $r_w$ and $H_l$ to have 4 levels each, and then creating a new qualitative variable $t$ whose 16 levels represent discrete combinations of those two inputs, as shown in Figure \ref{fig:borehole-true}. The total Sobol sensitivity indices \cite{sobol2001global} of the original numerical inputs $r_w$ and  $H_l$ are 0.86 and 0.05, respectively. Therefore, $r_w$ is much more important than $H_l$, and we would hope that the (representative) latent space for $t$ reflects this. In particular, we would hope that the levels corresponding to the same $r_w$ value (for e.g. levels 1,2,3, and 4) to be closer to each other on average than those with different $r_w$ values.  We consider two different training set sizes of $n=32$ and $n=64$ observations, corresponding to having 2 and 4 observations per each level. The training sets are generated using a two-step design of experiments (DoE) approach. First, the quantitative variables are generated using Latin hypercube sampling. Then, the qualitative factor levels of each data point are assigned using random stratified sampling, with stratification on the levels to ensure that each level occurs for the same number of training observations.

The estimated representative LV spaces for the two training sets are shown in Figures \ref{fig:borehole-32}
and \ref{fig:borehole-64}, respectively. In both cases, the $z_1$ LV corresponds roughly to $H_l$, and the $z_2$ corresponds fairly closely to $r_w$, more so for the set of 64 training observations. Moreover, levels with the same $r_w$ value are closer to each other than those with different $r_w$ values. Note that the distances between different levels in the (representative) LV space are larger in the case of the training set with 32 observations, and the agreement between the estimated LV space and the true LV space  is not as tight as for the case of 64 training observations. This is likely due to larger estimation uncertainty with just 2 observations per level, and hence, greater discrepancies between the different posterior LV samples. Also, the effect of $H_l$ is not very clear for the smaller training set, since the orderings of the $H_l$ levels are different  at different $r_w$ values. However, with the larger training set in Figure \ref{fig:borehole-64}, the effects of both $r_w$ and $H_l$ are more clear. In fact, we can conclude that $z_1$ approximately represents the effects of $H_l$, while the $z_2$ approximately represent the effects of $r_w$.

The above example shows that LVGP with Bayesian inference can offer useful interpretations even with small training datasets. As we collect more data, we can draw more definitive conclusions from the representative LV space.

\section{Scaling fully Bayesian inference for LVGPs using sparse approximations} \label{sec:sparse-lvgp}
GPs, and LVGPs by extension, do not scale well with $N$. The computational requirements scale as $O\smb{N^3}$ time and as $O\smb{N^2}$ memory. In addition, LVGPs have many more parameters to estimate than regular GPs, as the latent variables must also be estimated. Since fully Bayesian inference typically requires many more likelihood evaluations than does maximum likelihood estimation, approximations may be needed for certain applications with larger datasets. In this section, we consider sparse approximations to the GP likelihood \eqref{eq:likelihood} using inducing points. These methods lower the computational costs to $O\smb{NM^2}$ time and $O\smb{NM}$ memory, where $M<<N$ is a pre-defined number of inducing points. We also address a few challenging issues that arise in defining inducing points for the qualitative inputs. 
% Stochastic variational approximations \cite{hensman2015mcmc,wang2022scalable} can further reduce the computational costs to $O\smb{M^3}$ time and $O\smb{M^2}$ memory. However, they need to be used with MCMC algorithms that allow likelihood evaluations on subsets of data to enable this computational efficiency. This is beyond the scope of the current work. 

% These methods lower the computational costs to $O\smb{NM^2}$ time and $O\smb{NM}$ memory, where $M<<N$ is a pre-defined number of inducing points. These approximate models can then be used with standard MCMC algorithms.  

Inducing point methods augment the model with $M$ inducing points (also referred to as pseudo-inputs) $\{\mathbf{u}_1,\ldots,\mathbf{u}_M\}$ and their corresponding function values $\mathbf{f}_u = \left[f(\mathbf{u}_1),\ldots,f(\mathbf{u}_M)\right]$. A common assumption behind all inducing point methods is that the function value $f\smb{\mathbf{w}_*}$ at a test location $\mathbf{w}_*$ and the function values at the training locations $\mathbf{f} =\left[f(\mathbf{w}_1),\ldots,f(\mathbf{w}_N)\right] $ are conditionally independent given $\mathbf{f}_u$. Under this assumption, we use the standard approximation \cite{quinonero2005unifying,Titsias2009VariationalLO}
\begin{equation}
    p\smb{f\smb{\mathbf{w}_*},\mathbf{f}} \approx q\smb{f\smb{\mathbf{w}_*},\mathbf{f}} = \int   q\smb{f\smb{\mathbf{w}_*}|\mathbf{f}_u}q\smb{\mathbf{f}|\mathbf{f}_u} p\smb{\mathbf{f}_u} d\mathbf{f}_u,
    \label{eq:inducing-point-approx}
\end{equation}
\noindent where the $q\smb{\cdot}$ are some approximating distributions. The function values $\mathbf{f}_u$ are marginalized analytically in the right-hand side of \eqref{eq:inducing-point-approx}, leaving the inducing points $\mathbf{u}_1,\ldots,\mathbf{u}_M$ as the only additional parameters.
We consider two such methods - the fully independent training conditional (FITC) \cite{snelson2005sparse}, and the variational free energy (VFE) \cite{Titsias2009VariationalLO} methods. The two methods use slightly different approximating distributions $q$ and likelihood objectives but have quite different performances in practice. See \cite{bauer2016understanding} for more details.

\begin{figure}[tbhp]
    \centering
     \subfloat[Restricted to mapped values]{\label{fig:inducing-a}\includegraphics[width=0.36\textwidth]{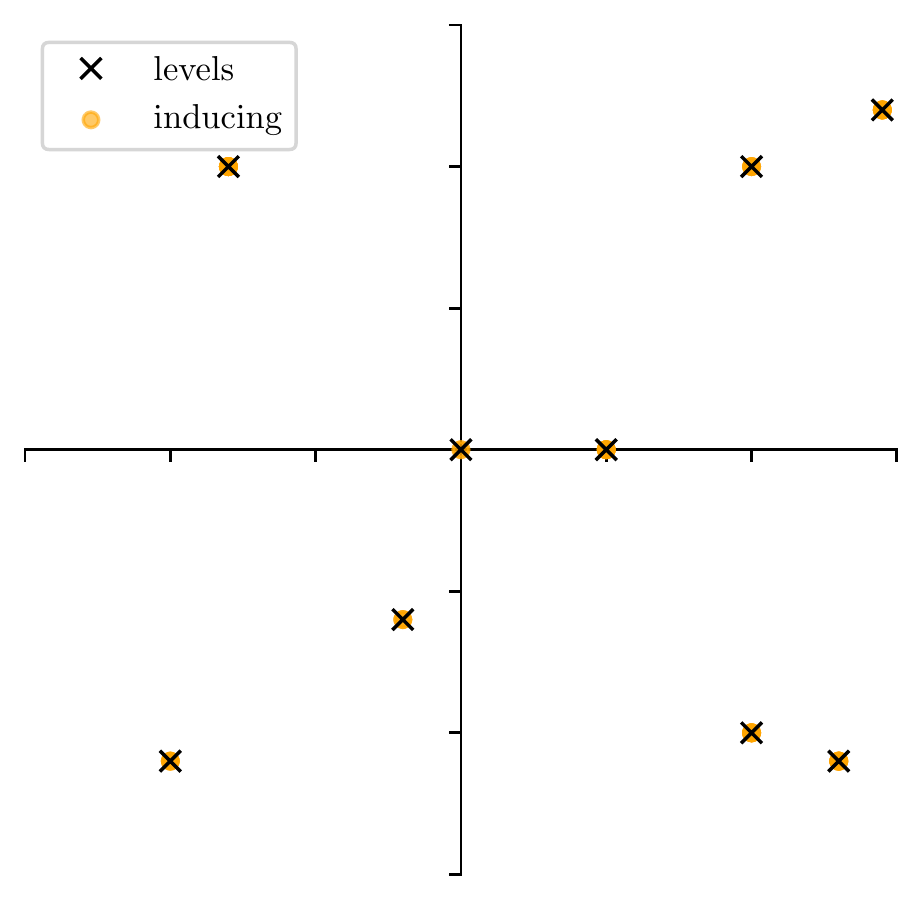}}
    \qquad\qquad
    \subfloat[Convex hull relaxation using \eqref{eq:inducing-convex}]{\label{fig:inducing-b}\includegraphics[width=0.36\textwidth]{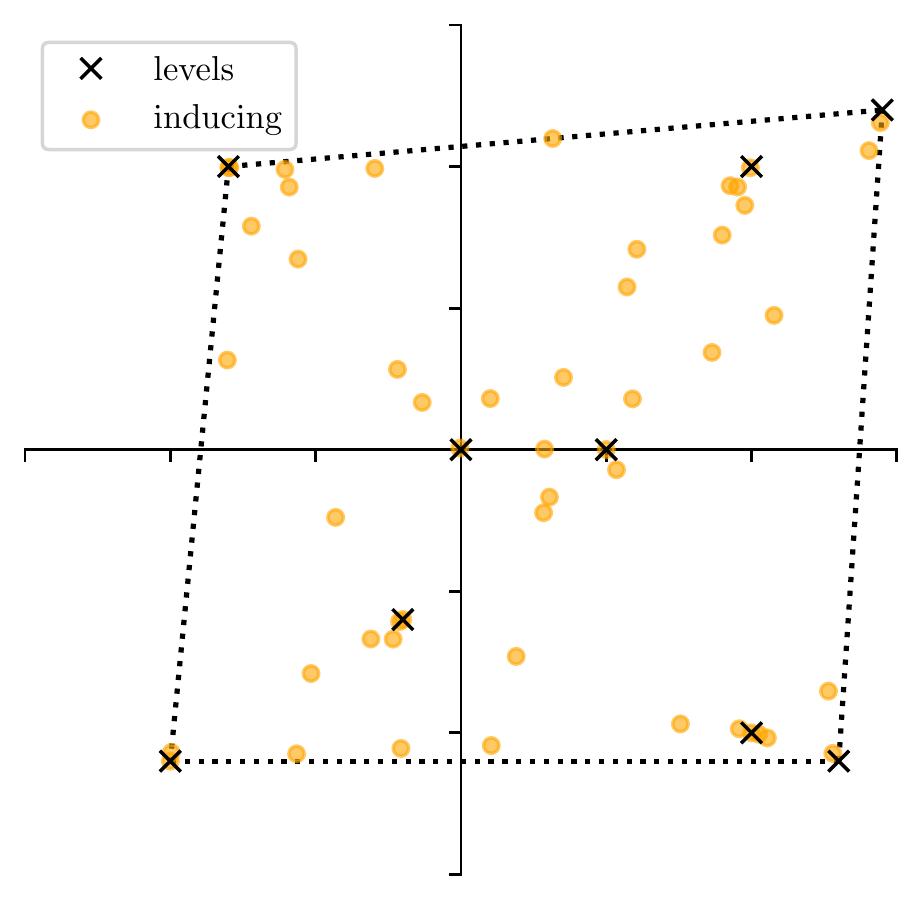}}
    \caption{Inducing points for a qualitative variable in its corresponding LV space}
    \label{fig:inducing-point-demo}
\end{figure}

These inducing point methods were developed for numerical inputs, for which the inducing points are continuous parameters. However, the domain for qualitative inputs is a finite set of values, which requires inference techniques over a  mixed-variable parameter space. To avoid difficulties associated with such inference, we first define the inducing points for a qualitative variable $t$ in its corresponding latent variable space. We then relax the constraint that they belong to the finite set of the mapped LV values $\{\mathbf{z}\smb{1},\ldots,\mathbf{z}\smb{L}\}$ (illustrated in Figure \ref{fig:inducing-a}), and allow them to lie in the convex hull of this set, as illustrated in Figure \ref{fig:inducing-b}. To ensure the inducing points lie in the convex hull, we represent the $m^\mathrm{th}$ inducing point location for $t$, denoted by $\mathbf{u}_m^{\smb{t}}$, as
\begin{equation}
    \mathbf{u}_m^{\smb{t}} = \sum_{\ell=1}^{L} \psi_{m\ell}^{\smb{t}}\;\mathbf{z}\smb{\ell} \quad m \in \{1,\ldots M\},
    \label{eq:inducing-convex}
\end{equation}
\noindent where $\{\psi_{m1}^{\smb{t}},\ldots,\psi_{mL}^{\smb{t}}\}$ are weight parameters with $0\leq\psi_{m\ell}^{\smb{t}}\leq 1$, and $\sum_{\ell=1}^{L}\psi_{m\ell}^{\smb{t}} = 1$. These can be further expressed as a function of $L-1$ simple bound-constrained parameters (see \cite{betancourt2012cruising}, e.g.). If the mapped LV values are fixed, $\mathbf{u}_m^{\smb{t}}$ could potentially be represented using fewer weight parameters, i.e., those corresponding to the levels whose mapped values make up the vertices of the convex hull. However, the mapped LV values must be estimated from data in practice. The subset of levels whose mapped values make up the  vertices of the convex hull can repeatedly change during optimization which complicates the joint estimation of the inducing points and the LVs using this sparser parameterization.  In contrast, the formulation \eqref{eq:inducing-convex} allows for more convenient joint optimization of the inducing points and the LVs.

For tractable fully Bayesian inference over these sparse models, we perform fully Bayesian inference only over the LVs and other LVGP hyperparameters and we fix the parameters for the inducing points (of both the numerical and the qualitative variables) to their maximum likelihood estimates. 
% To avoid this, in \cite{wang2022scalable}, where a stochastic variational approximation to the LVGP model is used, the inducing points for a qualitative variable $t$ are defined as unconstrained continuous parameters in its corresponding LV space. However, we find that with the two sparse approximations, the joint optimization over inducing points and latent variables with this scheme is extremely sensitive to initialization. With this scheme, we find that the inducing points do not sufficiently cover the mapped LV values as illustrated in Figure ...  With fully Bayesian inference, fixing them to the MAP values often resulted in poor performance. 

%% file: empirical.tex
\section{Numerical studies} \label{sec:empirical}

In this section, we present numerical studies to compare plug-in inference with MAP estimates against fully Bayesian inference for the LVGP. As in \cite{zhang2019lvgp}, we will use a 2-D latent space for all the case studies. For each case study, we compare the methods using two metrics. The prediction quality of the methods is quantified as the relative root mean-squared error (RRMSE) of their predictions over $N$ test points:
\begin{equation}
    \mathrm{RRMSE} = \sqrt{
    \frac{\sumObs{i}{N} \smb{y_i-\widehat{y}_i}^2}{\sumObs{i}{N} \smb{y_i-\overline{y}}^2}
},
\end{equation}

\noindent where $y_i$ and $\widehat{y}_i$ denote the true and predicted values respectively for the $i^\mathrm{th}$ test sample, and $\overline{y}$ is the average across the $N$ true test observations. To assess the quality of UQ,  we use the negatively oriented mean interval score (MIS) \cite{gneiting2007strictly} for the 95\% (central) prediction intervals. For an observation $y_i$, the interval score for a 95\% prediction interval $\left[l_i,u_i\right]$  is given by 
\begin{equation}
\mathrm{IS}\smb{y_i,\left[l_i,u_i\right]} = \smb{u_i - l_i} + \frac{2}{0.05}\smb{l_i-y_i}_+ + \frac{2}{0.05}\smb{y_i - u_i}_+,
\end{equation}

\noindent where $\smb{\cdot}_+$ denotes the positive part of the argument. The interval score penalizes confidence intervals that do not contain their corresponding observations. Since fully Bayesian inference accounts for the uncertainty in estimating the LVs while plugin inference does not, the confidence intervals obtained through fully Bayesian inference will be wider than those obtained through plugin inference, and therefore are less likely to incur this penalty. However, the MIS also penalizes overly wide confidence intervals. The MIS results in Sections \ref{sec:engg} and \ref{sec:matdes_pred} indicate that our Bayesian LVGP model results in superior uncertainty quantification by appropriately widening, but not overly widening, the confidence intervals. Finally, we also report the coverage probabilities for these 95\% prediction intervals. The Python code used for these experiments can be found at \texttt{\url{https://github.com/syerramilli/lvgp-bayes}}.

\subsection{Engineering models} \label{sec:engg}  

We first compare the performance of the two methodologies on a set of engineering models that are commonly used for assessing surrogate models with numerical inputs. In each case, we discretize two numerical inputs and then construct a new qualitative factor $t$ whose levels represent discrete combinations of those two numerical inputs. In addition to the borehole model discussed in Section \ref{sec:interpret}, we also consider the following models:

\paragraph*{OTL Circuit}
The midpoint voltage of a transformerless (OTL) circuit function is 
\begin{equation*}
    B\frac{\smb{V_{b1}+0.74}\smb{R_{c2}+9}}{B\smb{R_{c2}+9} + R_f} 
    + 11.35 \frac{R_f}{B\smb{R_{c2}+9}+R_f} + 
    0.74B\frac{R_f}{R_{c1}} \frac{R_{c2}+9}{B\smb{R_{c2}+9}+R_f},
\end{equation*}

\noindent where $V_{b1} = 12R_{b2}/\smb{R_{b1}+R_{b2}}$, and the 6 inputs are ($R_{b1},R_{b2},R_f,R_{c1},R_{c2},B$). Refer to \cite{otl-piston} for complete definitions and physical meanings of these inputs. We discretize $R_f$ and $B$ to have 6 and 3 levels respectively, and define the 18 levels of $t$ as the Cartesian product of the levels of $R_f$ and $B$.

\paragraph*{Piston} The resulting cycle time for a piston moving within a cylinder is
\begin{align*}
    2\pi\sqrt{\frac{M}{
    k+ S^2\frac{P_0V_0}{V^2}\frac{T}{T_o}}},
\end{align*}

\noindent where $V = \frac{S}{2k}\sqrt{A^2+4k\frac{P_0}{T_0}T}$, A = $P_0S + 19.62 M - \frac{kV_0}{S}$, and the 7 inputs are ($M,S,V_0,k,$ $P_o,$ $T,T_0$). Refer to \cite{otl-piston} for complete definitions and physical meanings of these inputs.
We discretize $P_0$ amd $k$ to have 4 and 5 levels respectively, and define the 20 levels of $t$ as the Cartesian product of the levels of $P_0$ amd $k$.

For each system, we randomly sample 1000 points over the input space as test data over which the metrics are computed. We compare the estimation methods for three different training set sizes, corresponding to having 2, 3, and 4 observations, respectively, per level of the qualitative variable $t$. For example, the training set sizes for the borehole function would be 32, 48 and 64 respectively. The training sets are generated using a two-step design of experiments (DoE) approach. First, the quantitative variables are generated using Latin hypercube sampling. Then, the qualitative factor levels of each data point assigned using random stratified sampling, with stratification on the levels to ensure that each level occurs for the same number of training observations. We perform the analysis for 25 replicates, each using different training sets with a different random Latin hypercube sample in the quantitative variables. 

\begin{figure}[tbhp]
    \centering
    \includegraphics[width=\textwidth]{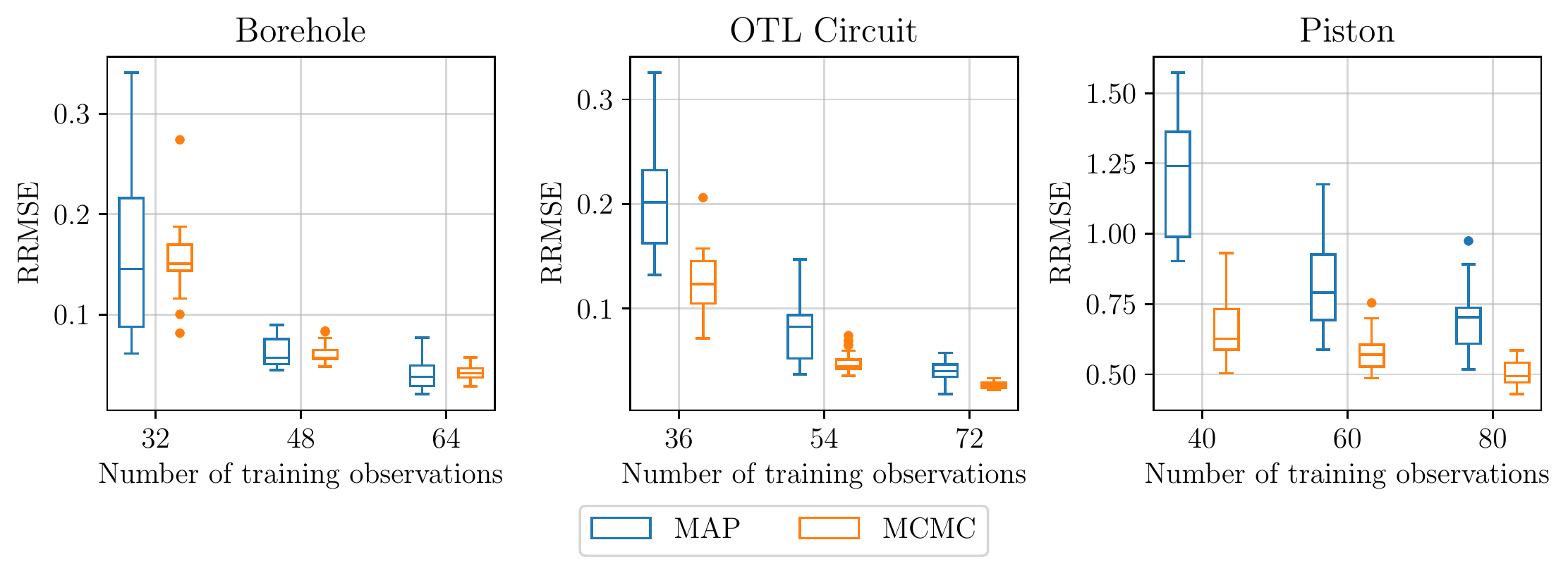}
    \caption{RRMSE  results across 25 replicates for the three different training set sizes for the three engineering models.}
    \label{fig:math-rrmse}
\end{figure}

The RRMSE results are shown in Figure \ref{fig:math-rrmse}. In all three systems, the prediction quality of fully Bayesian inference is either better than (sometimes substantially better) or comparable to that of plug-in inference. The fully Bayesian inference had slightly worse RRMSE on average only in the case of the borehole function with 32 training observations, but even in this case it resulted in better consistency across replicates (in Figure \ref{fig:math-rrmse}, its boxplot is narrower than that for the MAP approach). For most of the other situations in Figure \ref{fig:math-rrmse}, the fully Bayesian RRMSE is substantially better, with lower average RRMSE and better consistency (narrower box plots) across replicates. Regarding the consistency, even for the borehole function with 32 training observations, the fully Bayesian inference has better worse case performance, and is therefore more robust. 
\begin{figure}[tbhp]
    \centering
    \includegraphics[width=\linewidth]{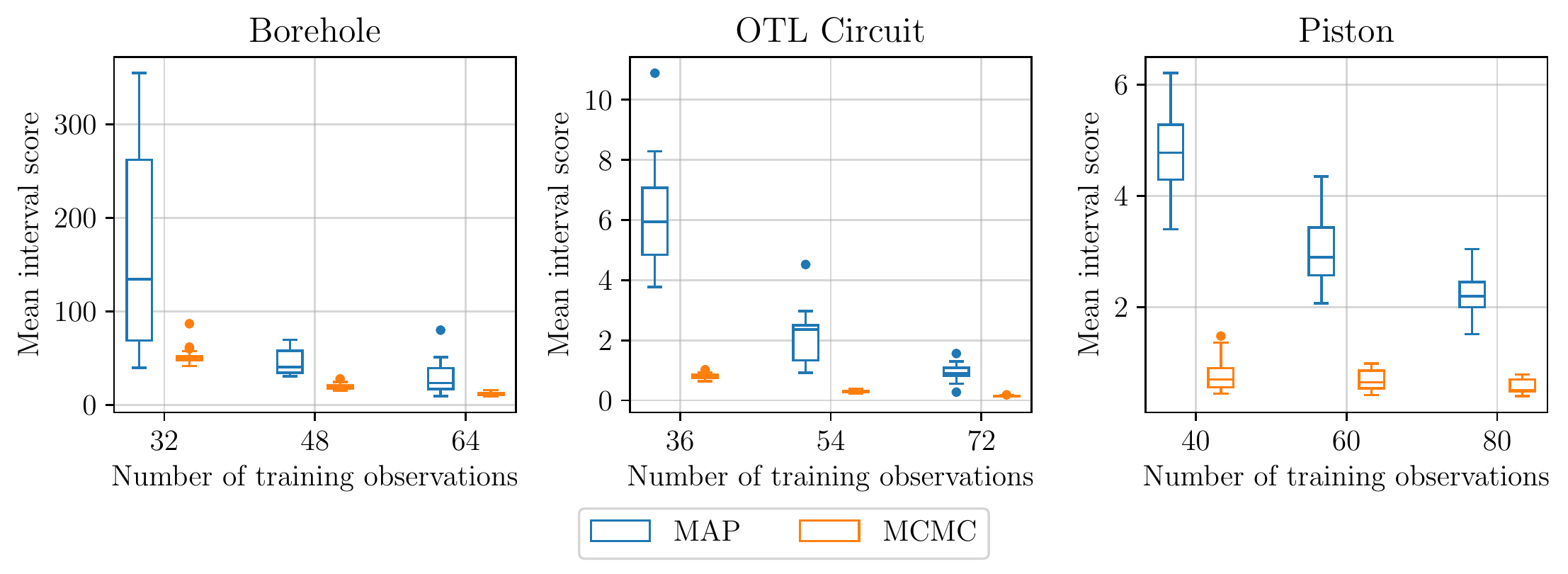}
    \caption{MIS results across 25 replicates for the three different training set sizes for the three engineering models.}
    \label{fig:math-mis}
\end{figure}

\begin{figure}[tbhp]
    \centering
    \includegraphics[width=\linewidth]{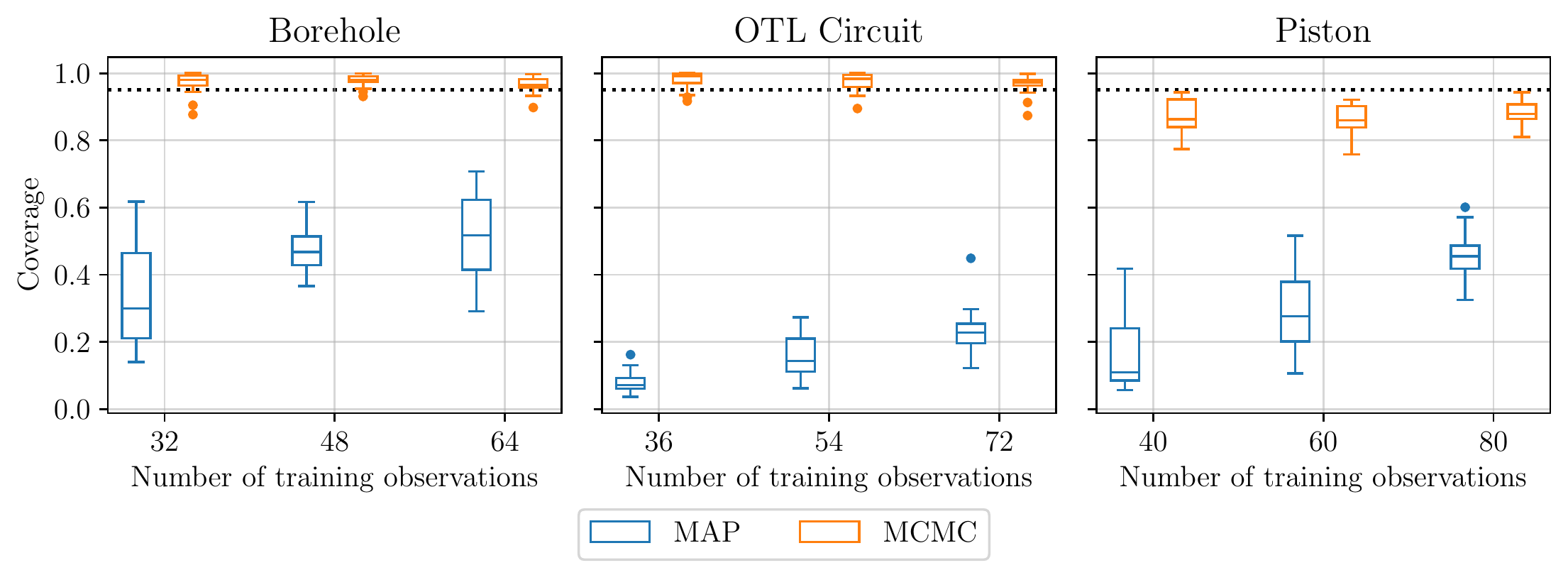}
    \caption{UQ coverage probabilities across 25 replicates for the three different training set sizes for the three engineering models.}
    \label{fig:math-cov}
\end{figure}

The corresponding MIS results are shown in Figure \ref{fig:math-mis}. Irrespective of training set size, there is a sizable improvement in the MIS from fully Bayesian inference. There is also much lower variance in the MIS across the different replicates, as is evident from the smaller interquartile ranges. The improvement in quality of UQ is expected, as there are many LVs to be estimated for each system, and hence the estimation uncertainty for the LVs will be large.  This is reflected in the coverage probabilities, shown in Figure \ref{fig:math-cov}. The MAP versions have poor coverage probabilities, especially for the OTL circuit function. While their coverage probabilities improve with increased training set sizes, they are still quite inaccurate. In contrast, with fully Bayesian inference, the models consistently have far better coverage probabilities irrespective of the training set size.

\subsection{Material design datasets} \label{sec:matdes_pred}

We now compare the estimation methodologies on the two different material design applications that motivated this work. In each case, we train independent LVGP models for multiple response properties of interest as a function of the atomic compositions of the materials, which constitute the purely qualitative inputs. In material design applications, the primary bottleneck is in obtaining physical measurements or first-principles calculations for the properties of interest, as these are typically expensive and time consuming \cite{himanen2019datadriven}. Therefore, it is important to develop cheap surrogate models that can learn efficiently from (relatively small) available datasets.

\subsubsection{Electronic properties of lacunar spinels} 
The first dataset consists of high-fidelity density functional theory (DFT) calculations for bandgaps and stabilities of 270 material compounds belonging to the lacunar spinel family $AM^aM_3^bQ_8$ \cite{wang2020featureless}. The $A$ atom has three levels $\{$Al, In, Ga$\}$, the $M^a$ atom has six levels $\{$V, Nb, Ta, Cr, Mo, W$\}$, the $M^b$ atom has five levels $\{$V, Nb, Ta, Mo, W$\}$, and the $Q$ atom has 3 levels $\{$S, Se, Te$\}$. In the original study \cite{wang2020featureless}, the authors used the LVGP with point estimates in a multi-objective LVGP-BO framework to find suitable metal-insulator transition materials in the lacunar spinel family. For this family of materials, choosing suitable numerical features for modeling the properties was challenging, as there was a lack of a clear understanding of the microscopic interactions governing the two properties. The LVGP methodology circumvented this issue by allowing modeling of the two properties as a function of the atomic compositions alone. 

Here, we are interested in assessing the change in model performance on small training sets through a fully Bayesian treatment of the LVGP hyperparameters. For this analysis, we generate training sets of sizes of 50, 75 and 100 through random sampling. For each training set of size $n$, we use the remaining $270-n$ observations as the test set over which the metrics are computed. We train and test the model across 25 replicates, where the replicates use different training and test splits. 

\begin{figure}[tbhp]
    \centering
    \includegraphics[width=0.8\textwidth]{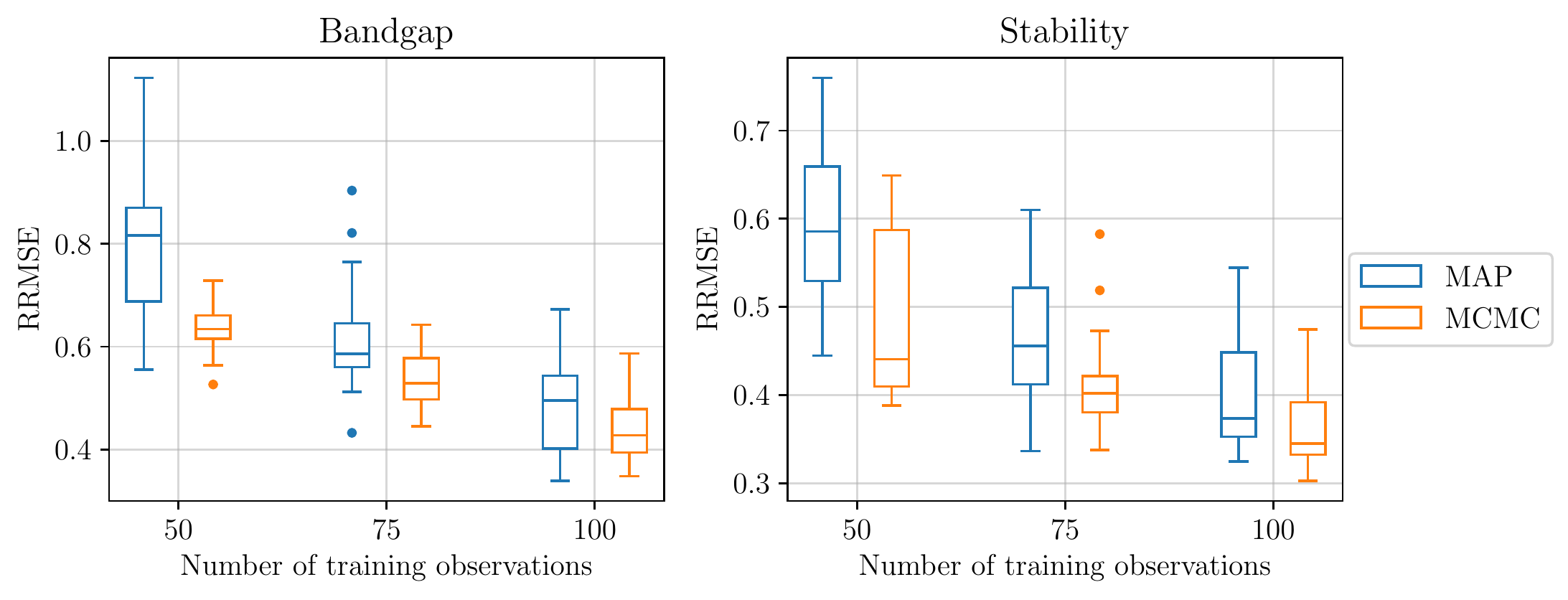}
    \caption{RRMSE results across 25 replicates for the two properties of interest in the lacunar spinal dataset. }
    \label{fig:spinels-rrmse}
\end{figure}

\begin{figure}[tbhp]
    \centering
    \includegraphics[width=0.8\textwidth]{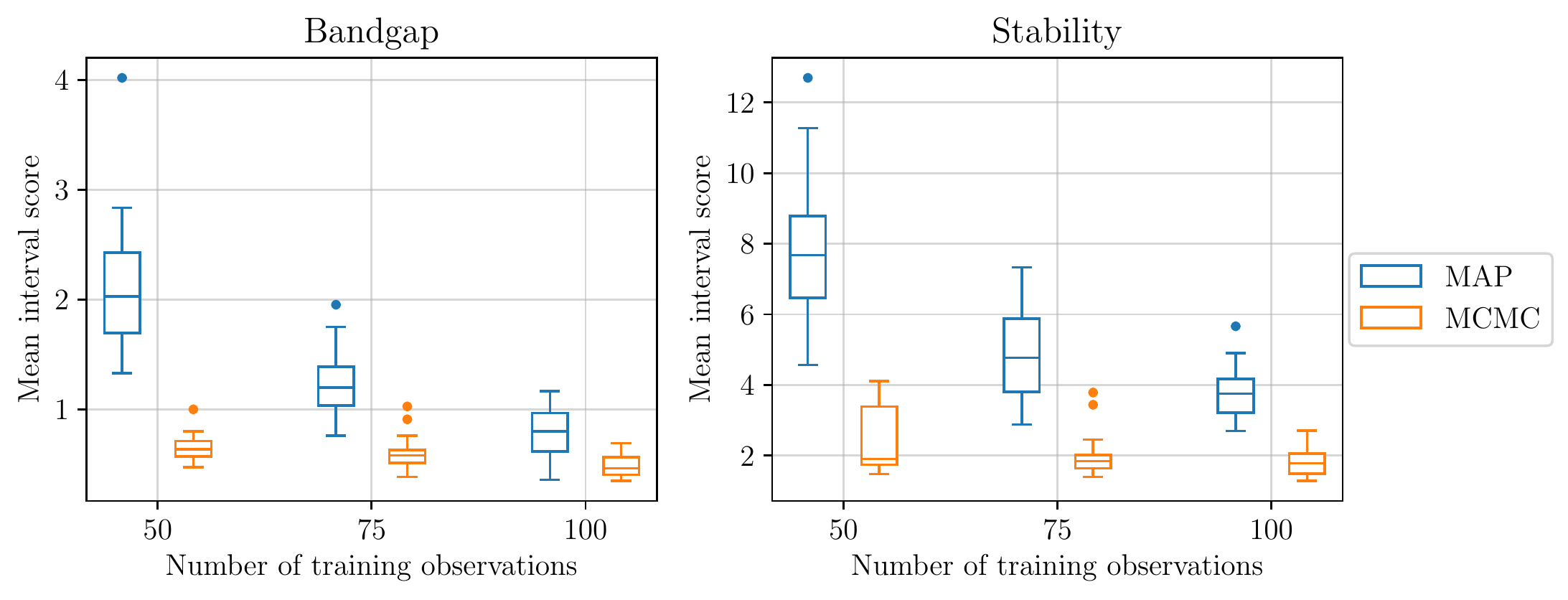}
    \caption{MIS results across 25 replicates for the two properties of interest in the lacunar spinal dataset.}
    \label{fig:spinels-mis}
\end{figure}

\begin{figure}[tbhp]
    \centering
    \includegraphics[width=0.8\textwidth]{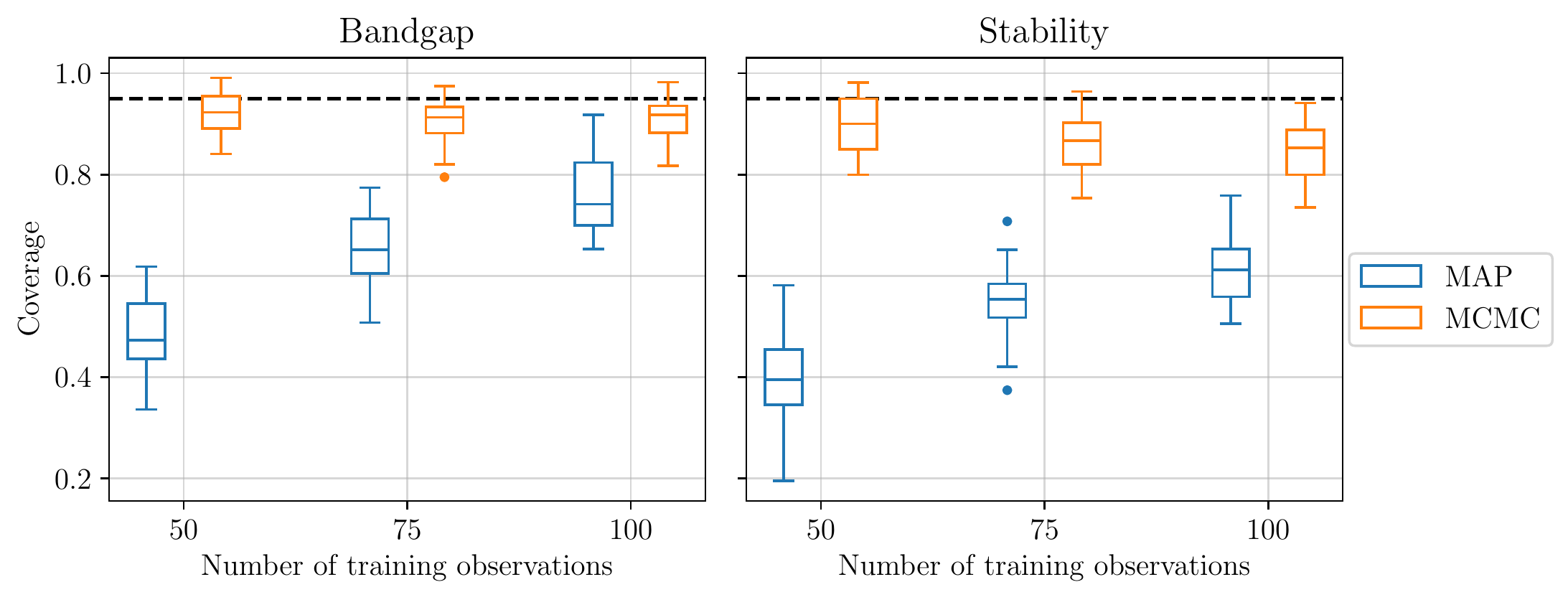}
    \caption{UQ coverage probabilities across 25 replicates for the two properties of interest in the lacunar spinel dataset.}
    \label{fig:spinels-cov}
\end{figure}

The RRMSE results are shown in Figure \ref{fig:spinels-rrmse}. For predicting both bandgap and stability, there is a large improvement from fully Bayesian inference irrespective of training set size. 
% For predicting stability, both approaches are comparable for a training set of size 20, while the performance of LVGP with fully Bayesian inference is much better for the larger training set sizes.
The corresponding MIS results are shown in Figure \ref{fig:spinels-mis}. The fully Bayesian inference provides a large improvement in the quality of UQ, and for all sample sizes considered. As in the previous case, this improvement in UQ is largely from much better coverage probabilities with fully Bayesian inference, as show in Figure \ref{fig:spinels-cov}. The extent of improvement in UQ  may seem a bit surprising given that none of the four qualitative inputs have many levels.  However, the \textit{total number} of LVs to be estimated across all four qualitative variables is still large and hence, there is a significant amount of estimation uncertainty. 

\subsubsection{Elastic moduli of $\text{M}_\text{2}$AX materials} \label{sec:bo-m2ax}

The second dataset is from another adaptive material exploration study \cite{balachandran2016adaptive} involving a combinatorial search of compounds belonging to the family of $\mathrm{M}_2\mathrm{AX}$ phases. The responses variables are the three different elastic moduli - Young's, shear, and bulk. The elastic moduli were estimated via DFT calculations. The M atom has ten levels $\{$Sc, Zr, Nb, Cr, Ti, Hf, V, W, Ta, Mo$\}$, the A atom has twelve levels $\{$Cd, Tl, In, Pb, Al, Ga, Sn, Ge, Si, As, P, S$\}$, and the X atom has two levels $\{$C, N$\}$. Among the 240 possible combinations, 17 were found to have negative elastic constants, and thus were not considered. 

In the original study \cite{balachandran2016adaptive}, the authors have considered GP modeling with a \textit{small} number of numerical descriptors for each qualitative factor, that were {expected} to have large effects on the response. They used seven descriptors in total: the s-, p-, and d-orbital radii for the M atom, and the s-, and p-orbital radii for the A and X atoms.  In the following, we compare the performance of the standard (for numerical inputs only) GP with the chosen descriptors against the LVGP with just the qualitative factors (i.e., the LVGP does not consider the numerical descriptors at all) for each of the three responses. In addition, we also compare the impact of fully Bayesian inference on the performance of both models. Since the X atom has only two levels, it makes no difference whether it is treated as a numerical descriptor or as a qualitative input. Therefore, we treat X as a binary numerical variable for both models. For analysis, 100 of the 223 data points are randomly selected as training data and the rest are used as the test data. We repeated this procedure for 25 replicates, where on each replicate we choose different random subsets to serve as training and test data. In all the 25 different training sets, there were no levels with missing observations.

\begin{figure}[tbhp]
    \centering
     \includegraphics[width=0.85\textwidth]{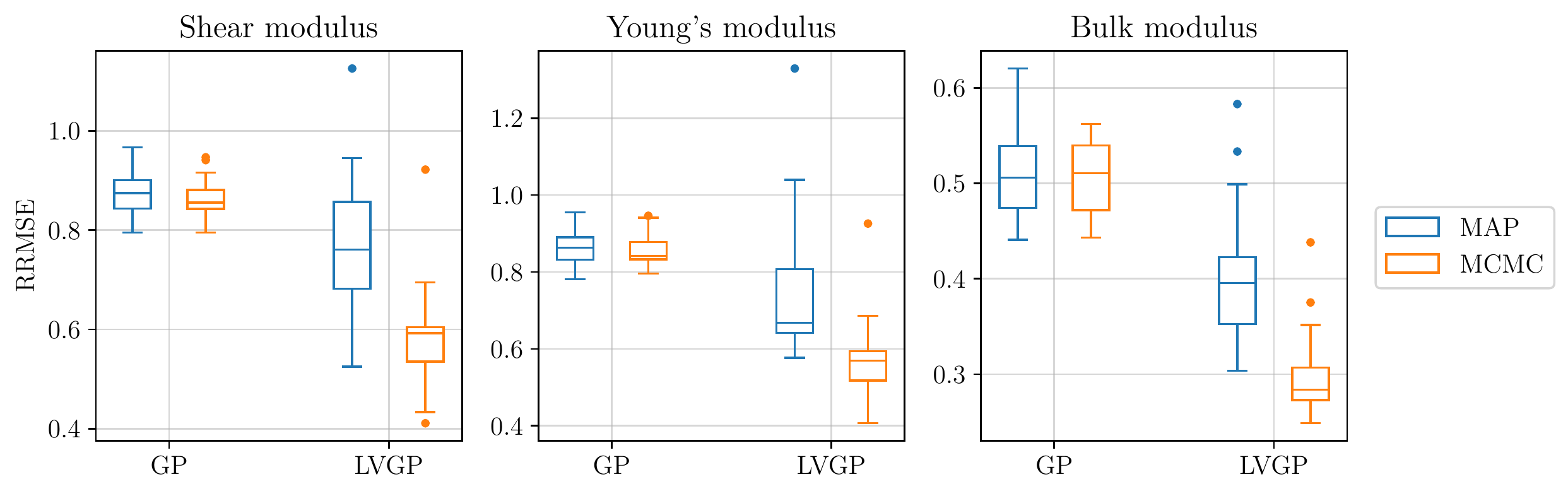}
    \caption{RRMSE metrics for the different models for the three responses in the $M_2AX$ dataset}
    \label{fig:m2ax-rrmse}
\end{figure}

The RRMSE results are shown in Figure \ref{fig:m2ax-rrmse}. Multiple observations that can be made from these results. Firstly, fully Bayesian inference results in a substantial improvement in the case of LVGP models, but offers much less improvement in the case of the GP models with only the numerical descriptors. This is perhaps due to the LVGP models having a much greater number of hyperparameters, and therefore a higher level of estimation uncertainty, in which case the MAP estimates may be misleading in the case of LVGPs. A second important observation from Figure \ref{fig:m2ax-rrmse} is that the GP models with the descriptors have consistently worse performance than the LVGP models (especially the fully Bayesian LVGP) on all three responses. This may be due to the numerical descriptors providing an insufficient representation of the effects of the qualitative variables. 

\begin{figure}[tbhp]
    \centering
     \includegraphics[width=0.85\textwidth]{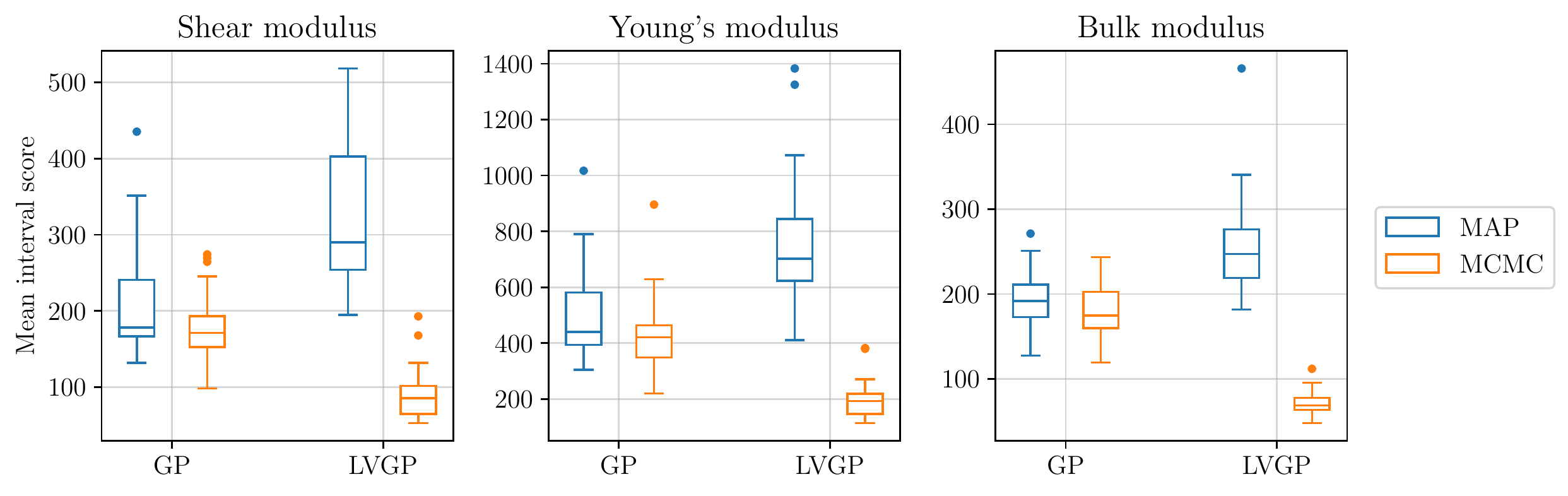}
    \caption{MIS metrics for the different models for the three responses in the $M_2AX$ dataset.}
    \label{fig:m2ax-mis}
\end{figure}

\begin{figure}[tbhp]
    \centering
     \includegraphics[width=0.85\textwidth]{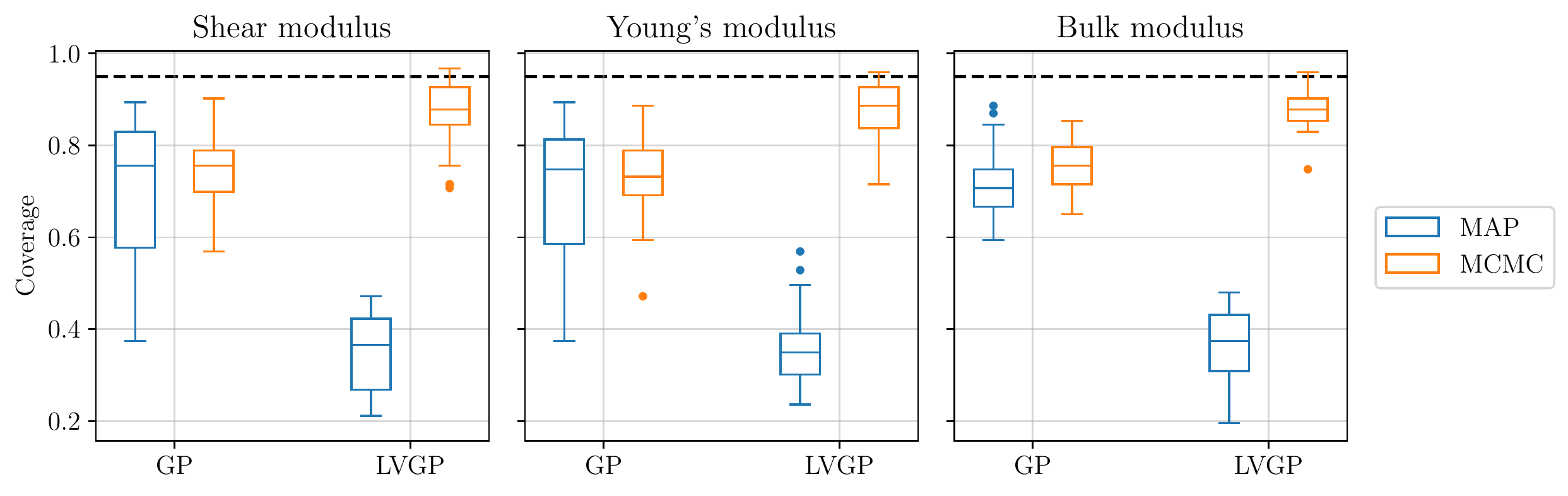}
    \caption{UQ coverage probabilities for the different models for the three responses in the $M_2AX$ dataset.}
    \label{fig:m2ax-cov}
\end{figure}

The corresponding MIS results are shown in the Figure \ref{fig:m2ax-mis}. As in the case of the RRMSE comparison, the fully Bayesian LVGP offers substantial improvement in UQ relative to its MAP version, whereas the fully Bayesian version of the GP model offers much less improvement over its MAP version. This intuitively makes sense, as LVGP models have more hyperparameters to estimate and hence the associated estimation uncertainty is much larger. Notice that despite having poorer predictive performance, the MAP version of the GP models have better UQ than the MAP version of the LVGP model. However, the fully Bayesian version of the LVGP model has substantially better UQ than any of the other methods (in addition to substantially better RRMSE). The trends observed in MIS are also reflected in the coverage probabilities, shown in Figure \ref{fig:m2ax-cov}.

\subsection{Formation energies of AB$\text{O}_\text{3}$ perovskites}

We now consider a dataset of DFT computed formation energies of AB$\mathrm{O}_{3}$ perovskites \cite{emery2017high}. These compounds are widely used for thermochemical water spitting applications. In these compounds, O is the oxygen atom, while A and B are variables that represent the cations that may occupy these two sites of the compound. There are 73 elements from the periodic table that can occupy either of these sites. The dataset contains a total of 5276 compounds for which formation energies are available. To test the approach, we randomly select 1000 observations as training data, and the remainder of the 5276 compounds are used as test data. We repeated this procedure for 25 replicates, where on each replicate we choose different random subsets to serve as training and test data. In each of the 25 different training sets, there were no levels with missing observations. For this application, fully Bayesian inference with all $N$ = 1000 observations was too prohibitively expensive to run across multiple replicates. Instead, we compare the MAP version of the LVGP model with all $N$ = 1000 observations (which we refer to as the exact MAP) versus both MAP and fully Bayesian versions of the two sparse LVGP approximations discussed in Section \ref{sec:sparse-lvgp}. For each sparse approximation, we set the number of inducing points to 50.

\begin{figure}[tbhp]
    \centering
    \subfloat[RRMSE results]{\label{fig:abo3-rrmse}\includegraphics[width=0.45\textwidth]{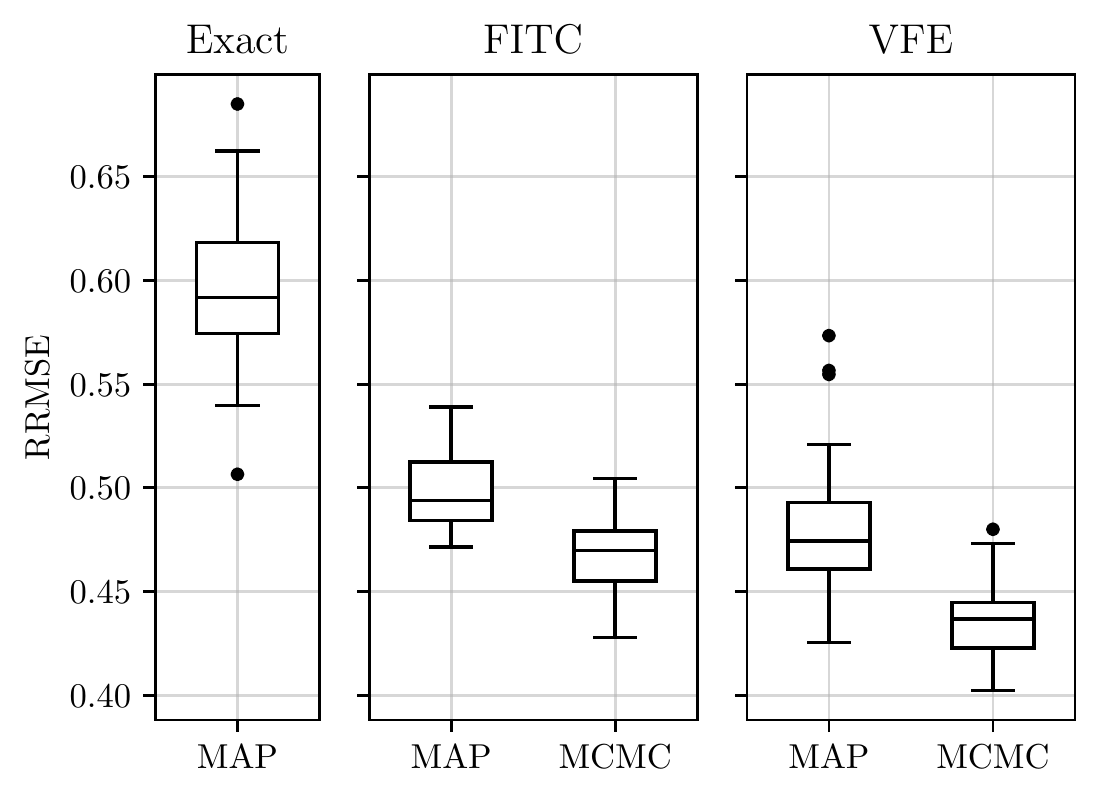}}
    \qquad
    \subfloat[MIS results (in log scale)]{\label{fig:abo3-mis}\includegraphics[width=0.45\textwidth]{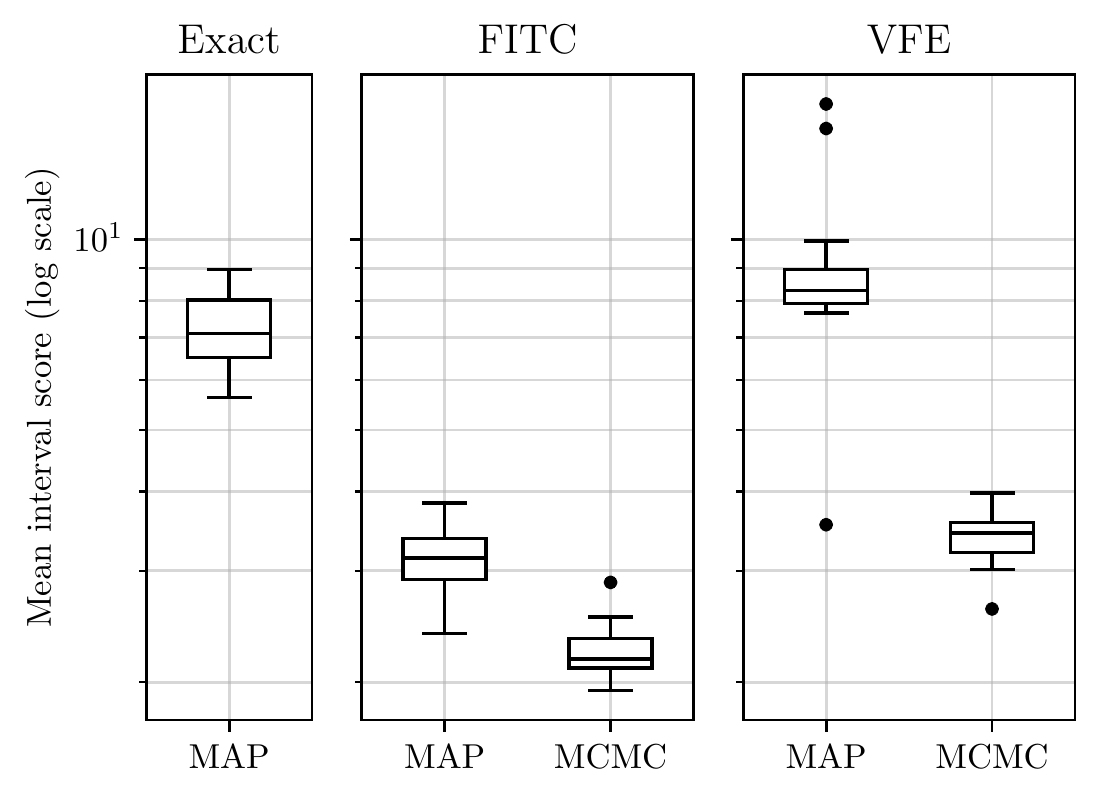}}
    \\
    \subfloat[Coverage probabilities]{\label{fig:abo3-cov}\includegraphics[width=0.45\textwidth]{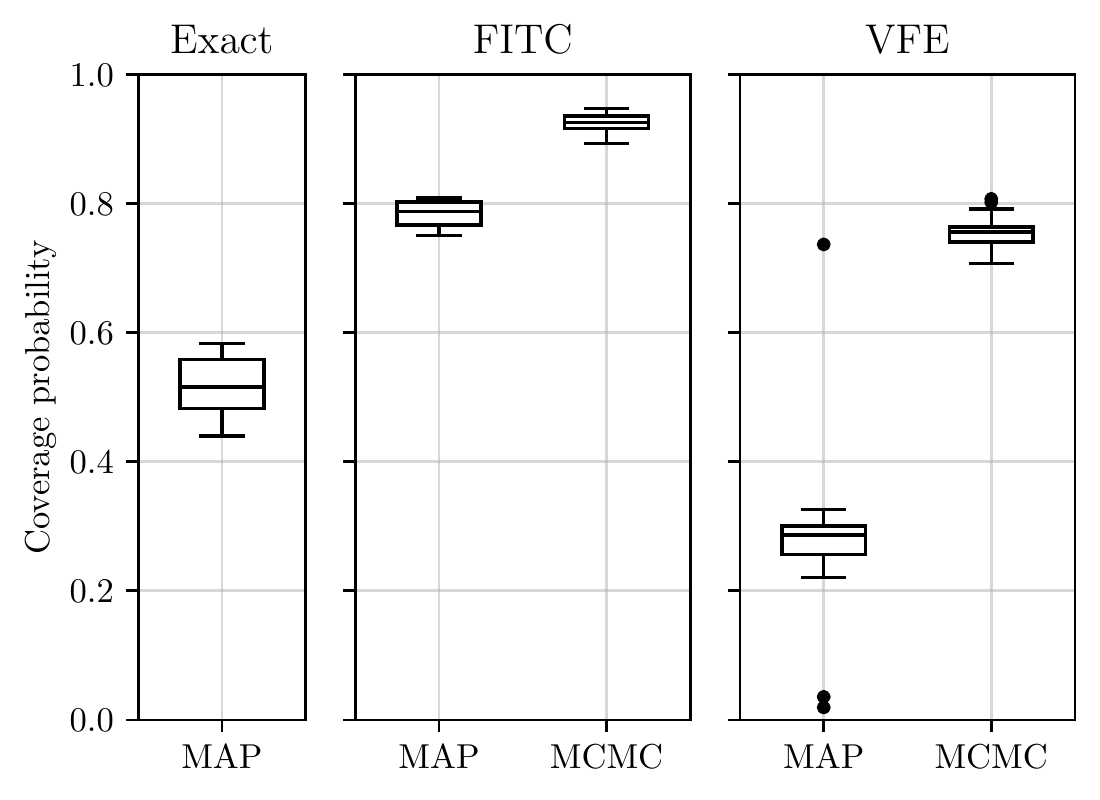}}
    \qquad
    \subfloat[Training times ]{\label{fig:abo3-time}\includegraphics[width=0.45\textwidth]{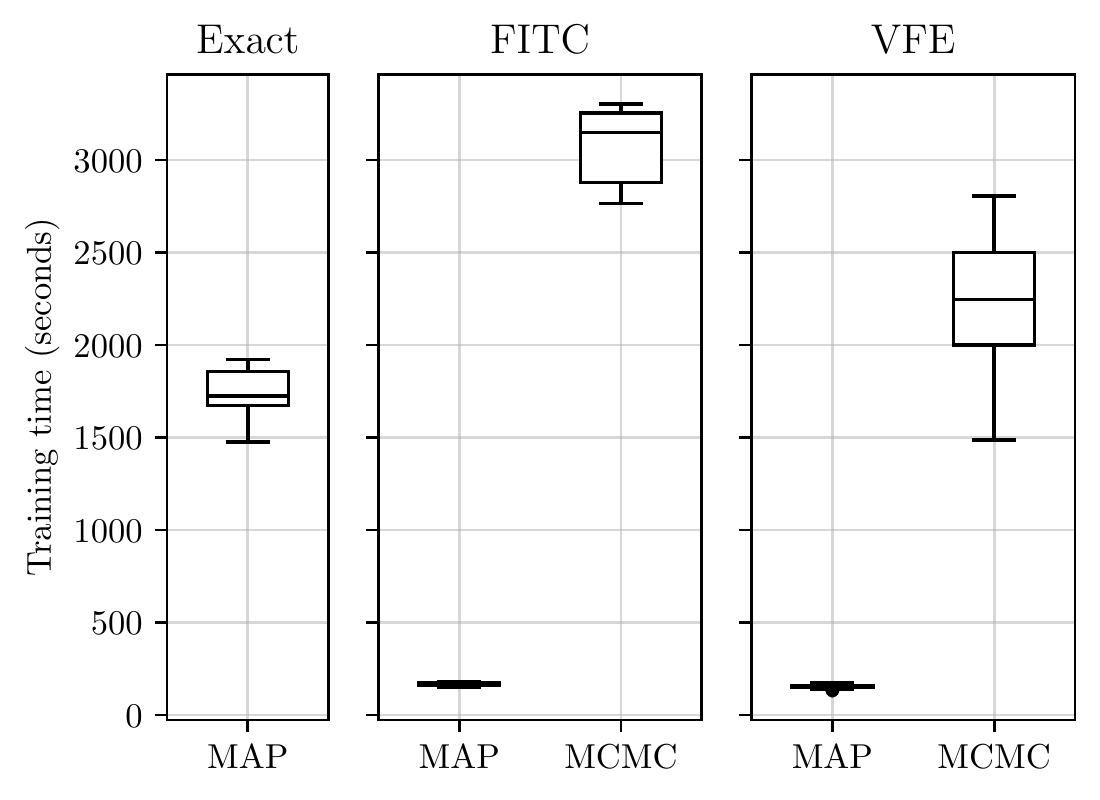}}
    \caption{Metrics for the different models for predicting formation energies of AB$\text{O}_3$ compounds.}
    \label{fig:abo3-results}
\end{figure}

The RRMSE results are shown in Figure \ref{fig:abo3-rrmse}. The first panel corresponds to the exact MAP LVGP, the second to the FITC approximation (MAP and fully Bayesian), and the third to the VFE approximation (MAP and fully Bayesian). Comparing the MAP versions, we can see that the sparse models have much better prediction accuracies than the exact MAP LVGP model, which may seem surprising. However, the optimization of the exact MAP LVGP model suffers from {stability issues} arising from computing the Cholesky decomposition of a {much larger} (covariance) matrix than that for the sparse LVGP models, which explains its poorer performance. The VFE model has slightly better prediction accuracy than the FITC model. For both models, there are significant improvements in prediction accuracy with fully Bayesian inference versus MAP.

The MIS results are shown in Figure \ref{fig:abo3-mis}. Comparing the MAP versions, the VFE model has worse UQ than the exact LVGP model. However, the FITC model has much better UQ than the exact LVGP model. With fully Bayesian inference, there are large improvements in the quality of UQ for both the sparse models. The relative improvement in UQ for the VFE model with the fully Bayesian inference is much larger than that for the FITC models. This can be explained by their respective coverage probabilities, which are shown in Figure \ref{fig:abo3-cov}. The coverage probability of the MAP FITC model is much larger than the of the MAP VFE model.

These results show that fully Bayesian inference for the LVs and the other hyperparameters results in improved performance even for larger training sets. However, these improvements come with a large increase in computation time. As shown in Figure \ref{fig:abo3-time}, the training time for fully Bayesian inference (which include the training time for obtaining the MAP estimates for the inducing points) is about an order of magnitude larger than that for obtaining the corresponding MAP estimates.

%% file: conclusions.tex
\section{Conclusions} \label{sec:conc}

In this work, we have developed a fully-Bayesian approach for LVGP modeling and conducted numerical studies comparing it to plug-in inference on a few engineering models and real datasets for material design problems. This involved developing prior distributions for the latent variables and developing an approach for interpreting the LVGP model with fully Bayesian inference. We have also developed approximations for scaling up LVGPs and for fully Bayesian inference for the LVGP hyperparameters using sparse inducing point methods. Through our numerical studies, we have observed substantial improvements in UQ from the fully Bayesian approach in every example that we considered. Moreover, and somewhat surprisingly, the predictive accuracy, as measured by the RRMSE, was often improved substantially and never significantly worsened. Finally, the LVGP with fully Bayesian inference is as  interpretable as the LVGP with point estimates using our proposed approach for constructing a single representative LV space. We, therefore, advocate for a fully Bayesian treatment of the LVGP hyperparameters, especially when one or more qualitative inputs have many levels.

\section*{Acknowledgements}
This work was supported by the Advanced Research Projects Agency-Energy (ARPA-E), U.S.\ Department of Energy, under Award Number DE-AR0001209. This research was also supported, in part, through the computational resources and staff contributions provided for the Quest high-performance computing facility at Northwestern University, which is jointly supported by the Office of the Provost, the Office for Research, and Northwestern University Information Technology.

%% file: supp.tex
\appendix 

\section{Sensitivity of results to the choice of hyperprior} \label{sec:app-hyperprior}

\begin{figure}[tbhp]
    \centering
    \includegraphics[width=0.7\textwidth]{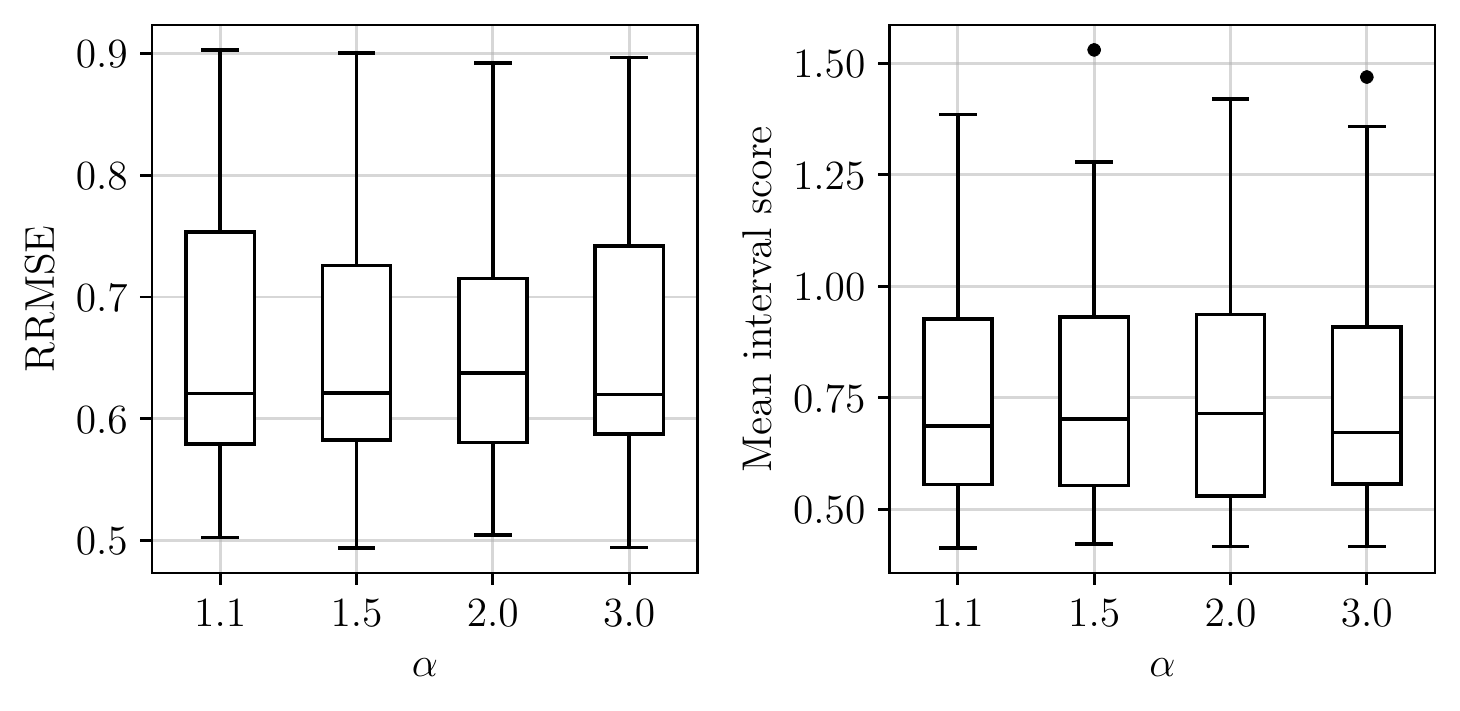}
    \caption{RRMSE and MIS results for the LVGP model with different $\alpha$ hyperparameters across 25 replicates for the Piston function with $N=40$ training observations}
    \label{fig:piston-40-alpha}
\end{figure}

\begin{figure}[H]
    \centering
    \includegraphics[width=0.7\textwidth]{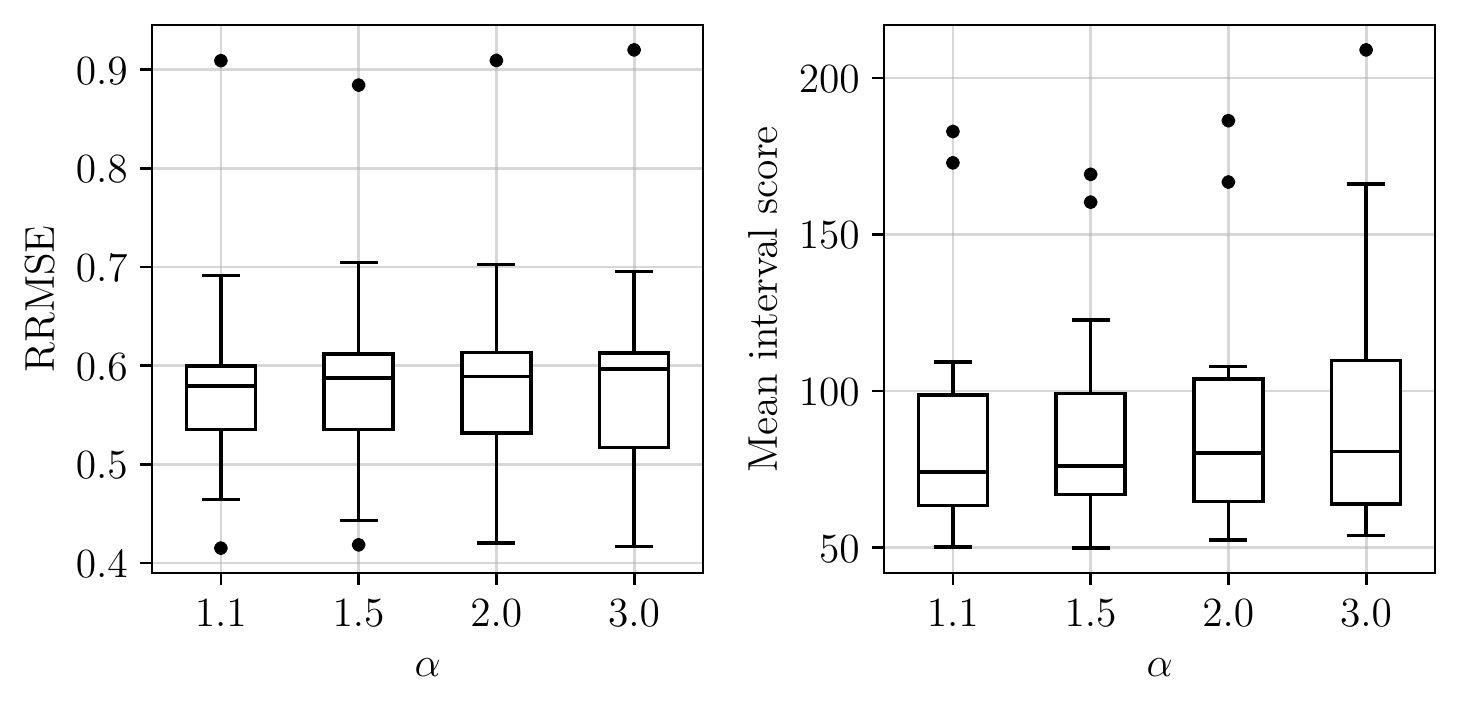}
    \caption{RRMSE and MIS results for the LVGP model with different $\alpha$ hyperparameters across 25 replicates for modeling Shear modulus}
    \label{fig:m2ax-shear-alpha}
\end{figure}

As mentioned in Section \ref{sec:priors}, we place a Gamma prior  on the precision hyperparameter $\gamma$, and constrain the concentration hyperparameter $\alpha > 1$ and set the rate hyperparameter $\beta = \alpha-1$, so that the mode of this prior will always be 1, irrespective of the value of $\alpha$. In this section, we test the sensitivity of the LVGP model to the value of $\alpha$ on a couple of cases in Section 5. Refer to Figures \ref{fig:piston-40-alpha} and \ref{fig:m2ax-shear-alpha}. We can see that there are no significant differences in prediction accuracy and uncertainty quantification across different values of $\alpha$ between 1.1 and 3.0.

\section{Training times}

In this section, we report the training times for both MAP and fully Bayesian inference for the examples discussed in Sections \ref{sec:engg} and \ref{sec:matdes_pred}. Note that we use different libraries for performing MAP estimation and fully Bayesian inference. We use GPyTorch \cite{gardener2018gpytorch}, PyTorch and Scipy for obtaining the MAP estimates, while we use JAX and NumPyro \cite{phan2019composable,bingham2019pyro} for running the No-U-Turn sampler \cite{hoffman2014no} algorithm. As such, they have different computational overheads, which can be significant for small training set sizes. 

\begin{figure}[tbhp]
    \centering
    \includegraphics[width=0.8\textwidth]{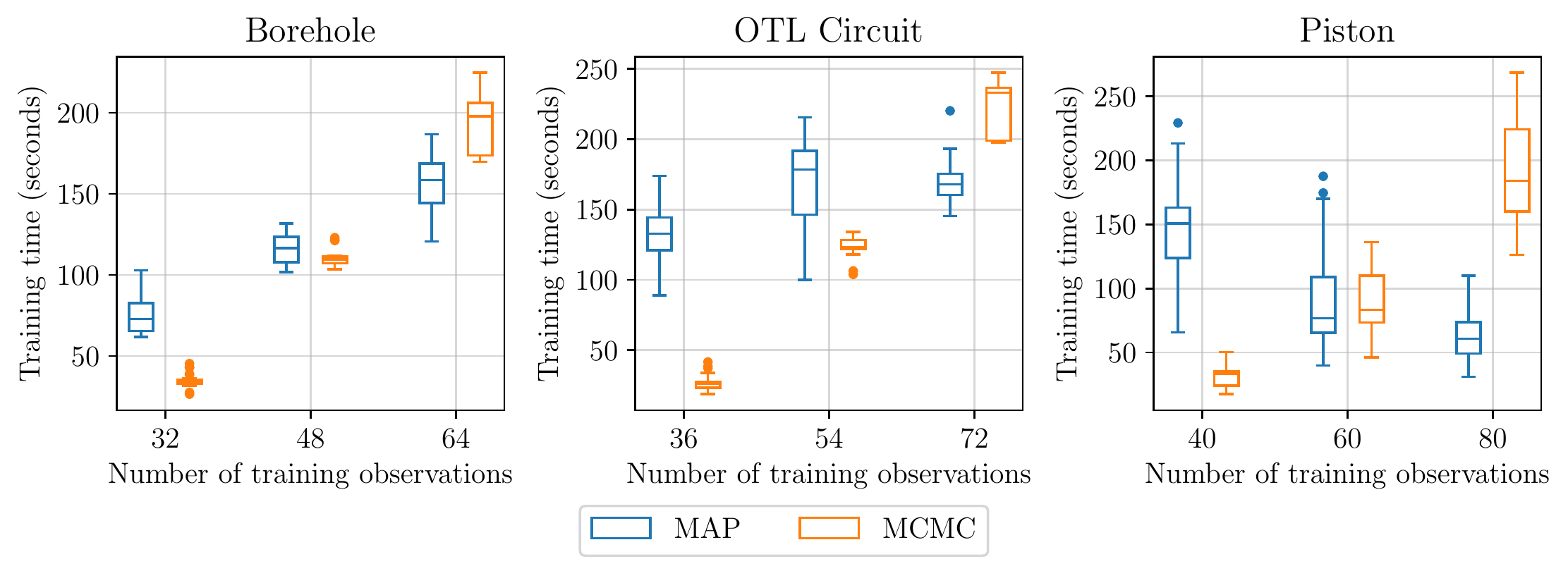}
    \caption{Training times across 25 replicates for the three different training set sizes for the three engineering models in Section 5.1}
    \label{fig:math-times}
\end{figure}

The training times for the engineering models discussed in Section 5.1 are shown in Figure \ref{fig:math-times}. For these models, fully Bayesian inference is actually faster than obtaining MAP estimates with smaller training set sizes, which could be due to a larger computational overhead with obtaining the MAP estimates. However, for the largest training set in each case, fully Bayesian inference is slower. Another odd result here is that for the piston model, the training times for the MAP models actually decreased with the number of training observations. On a closer look, we found that the total number of likelihood evaluations during optimization also decreased considerably with the number of training set observations, and this decrease more than offset the increase in the computational expense of each likelihood evaluation. 

\begin{figure}[H]
    \centering
    \includegraphics[width=0.7\textwidth]{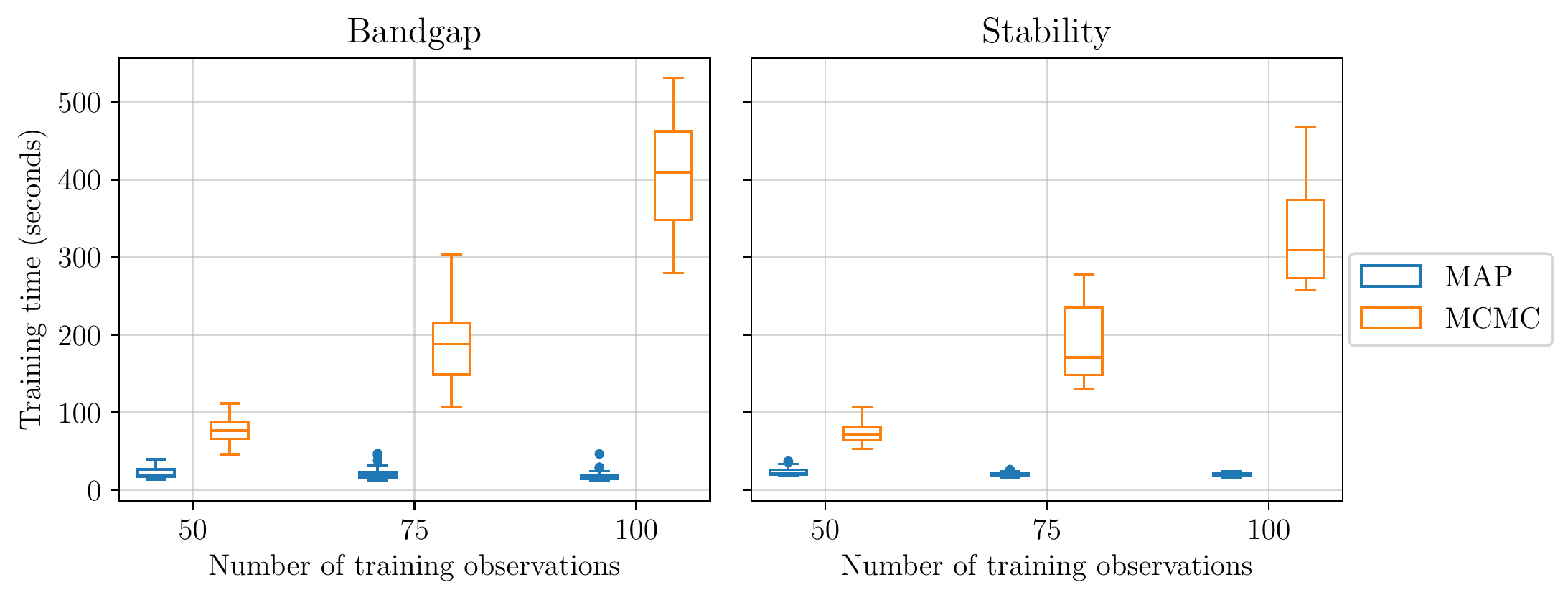}
    \caption{Training times across 25 replicates for the two properties of interest in the lacunar spinel dataset.}
    \label{fig:spinels-times}
\end{figure}

\begin{figure}[H]
    \centering
    \includegraphics[width=0.6\textwidth]{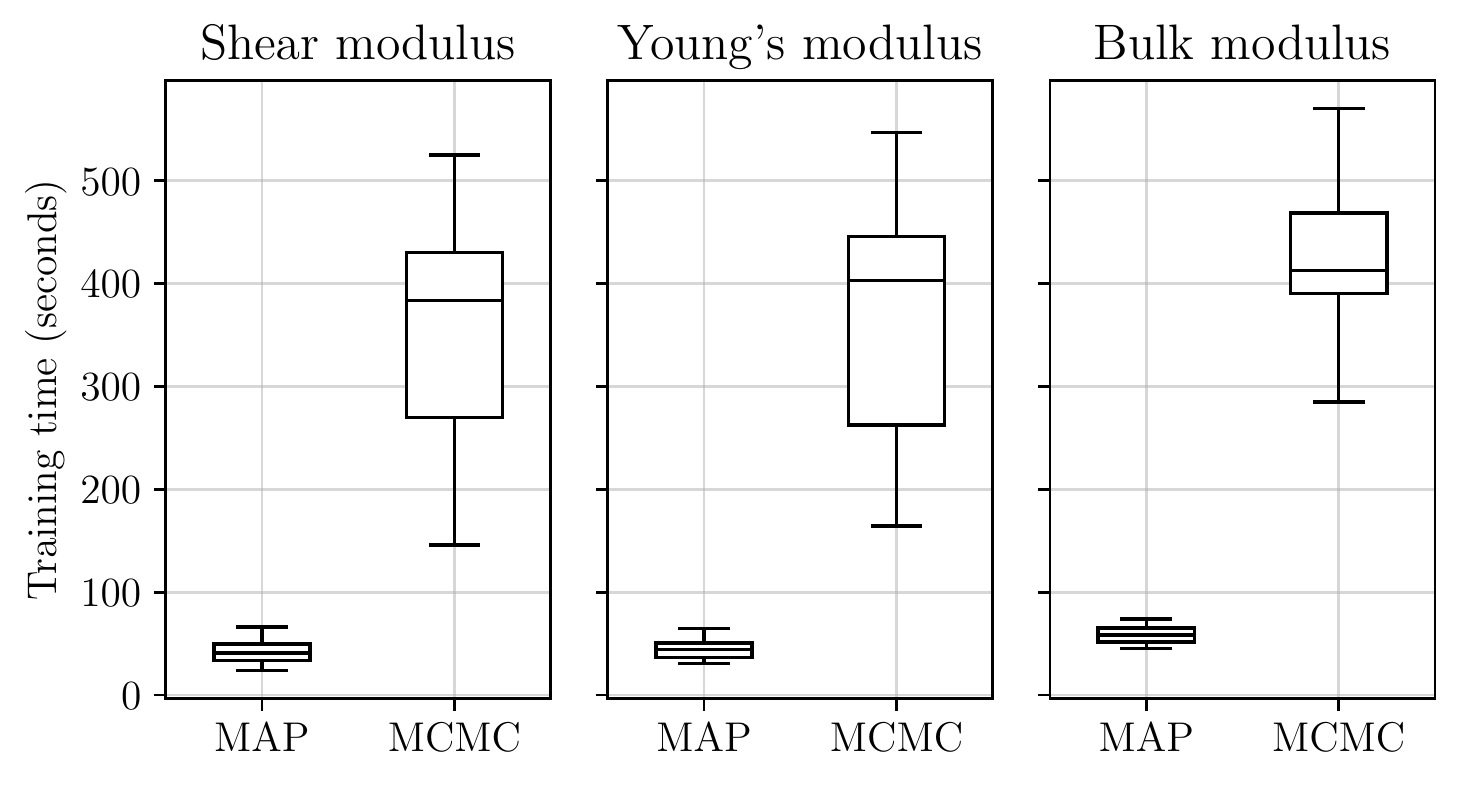}
    \caption{Training times across 25 replicates for the different LVGP models for the three responses in the $M_2AX$ dataset.}
    \label{fig:m2ax-times}
\end{figure}

The training times for the case of the lacunar spinels and the $M_2AX$ compounds are shown in Figures \ref{fig:spinels-times} and \ref{fig:m2ax-times}, respectively. For the $M_2AX$ compounds and the largest training set size for the case of the spinels, fully Bayesian inference is much more slower than MAP inference.

\section{Comparison of the representative latent space and the MAP latent space}

\begin{figure}[tbhp]
    \centering
    \includegraphics[width=0.7\textwidth]{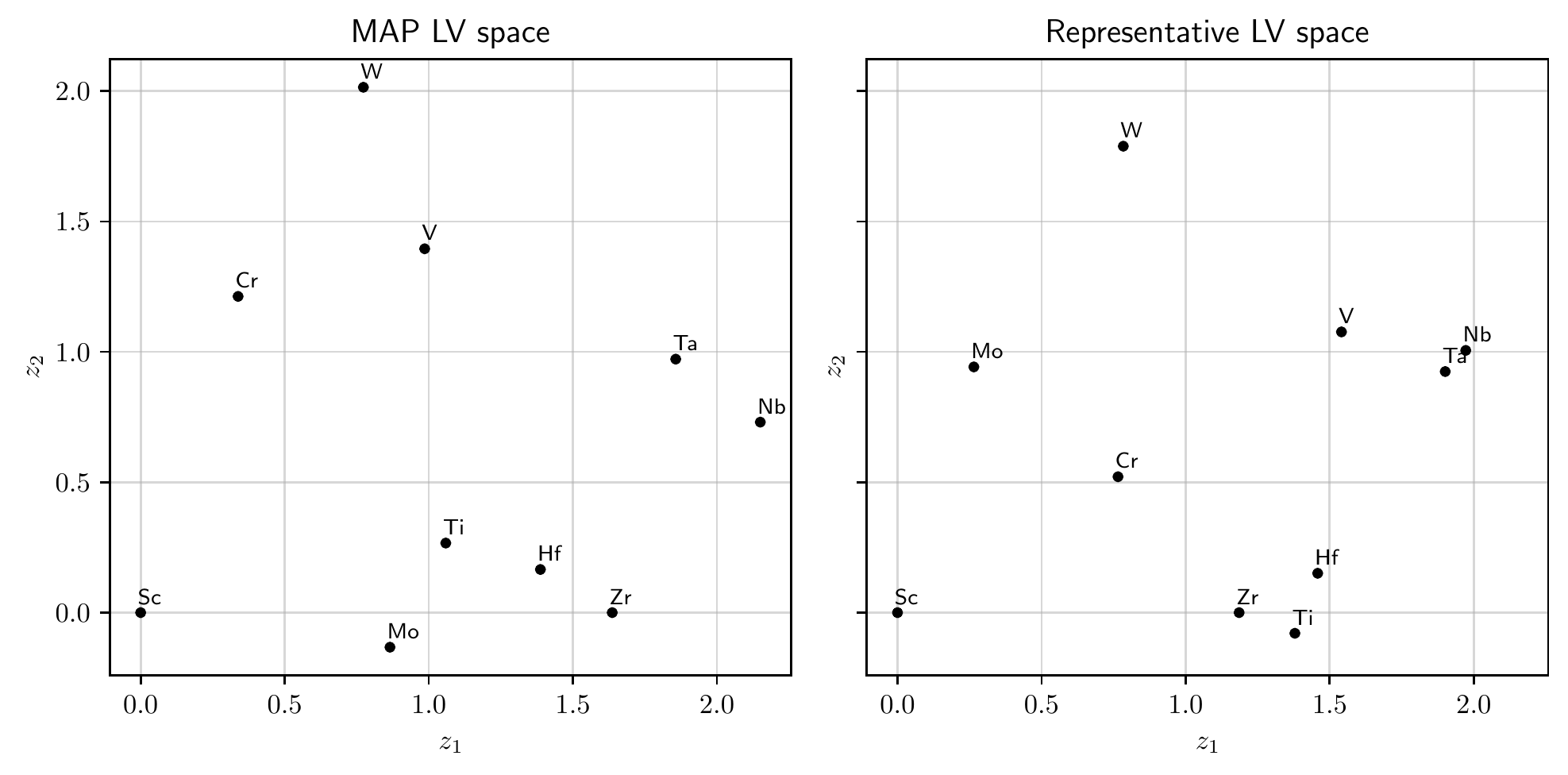}
    \caption{The latent space corresponding to the MAP estimate (left plot) versus the representative latent space (right plot) for the M-site element in the $M_2AX$ case study when modeling Shear modulus as a function of the composition}
    \label{fig:m2ax-rep-vs-map}
\end{figure}

In Section \ref{sec:interpret}, we proposed to construct a representative latent space to average out the sample-to-sample differences in the estimated LVs across each MCMC sample, and then use this representative latent space to interpret the effects of the qualitative variable. Alternatively, one might consider choosing the single latent sample corresponding to the MAP estimate for interpretation. These latent spaces can differ in terms of the positions of the levels in the latent space, and hence can result in slightly different interpretations. 
Consider the scenario in Figure \ref{fig:m2ax-rep-vs-map}, where we plot the latent space corresponding to the MAP estimate (left plot) and the representative latent space (right plot) for the M-site element when modeling the Shear modulus of $M_2AX$ materials. 
There are a few differences in the relative positions of the levels. For example, element Mo is close to elements Ti, Hf, and Zr in the MAP latent space, while it is much further away from them in the representative latent space. Aside from these differences, the two latent spaces largely agree with each other.

\section{Comparison with maximum likelihood estimates}

\begin{figure}[H]
    \centering
    \includegraphics[width=\textwidth]{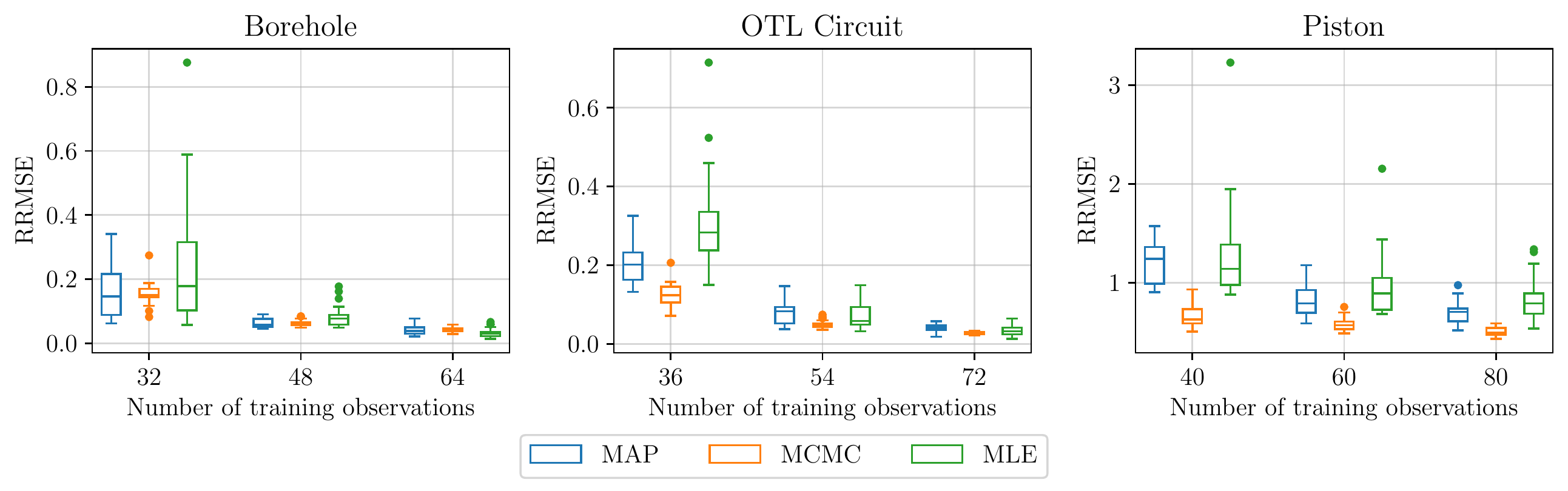}
    \caption{RRMSE  results across 25 replicates for the three different training set sizes for the three engineering models.}
    \label{fig:math-rrmse-mle}
\end{figure}

\begin{figure}[tbhp]
    \centering
    \includegraphics[width=\textwidth]{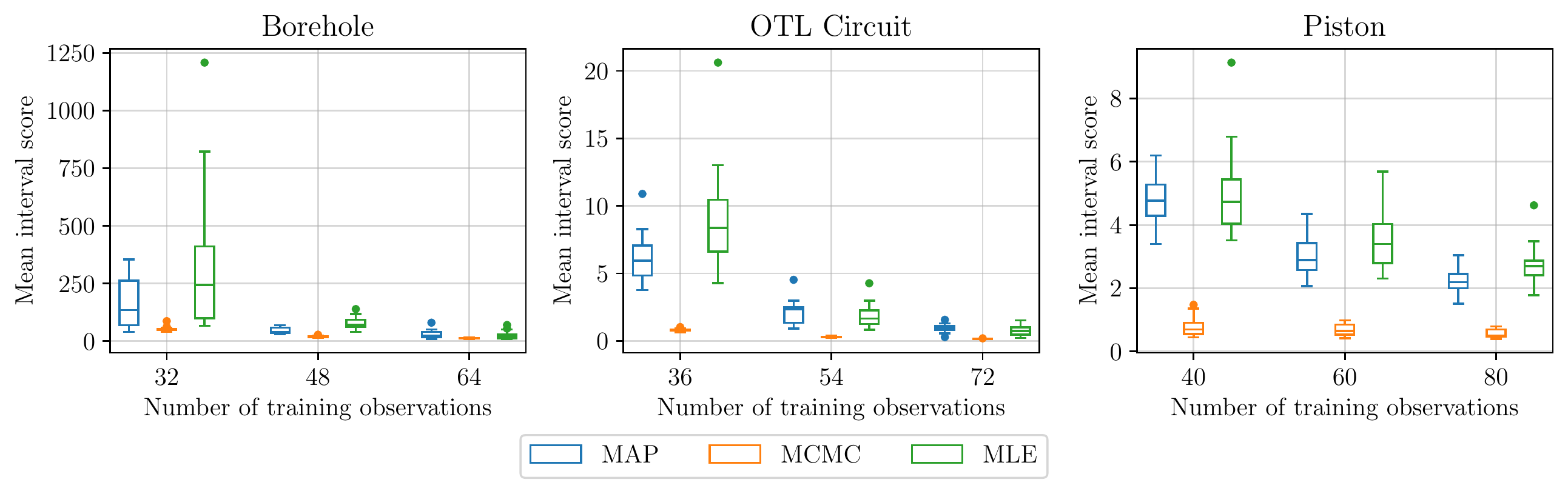}
    \caption{MIS results across 25 replicates for the three different training set sizes for the three engineering models.}
    \label{fig:math-mis-mle}
\end{figure}

\begin{figure}[tbhp]
    \centering
    \includegraphics[width=0.8\textwidth]{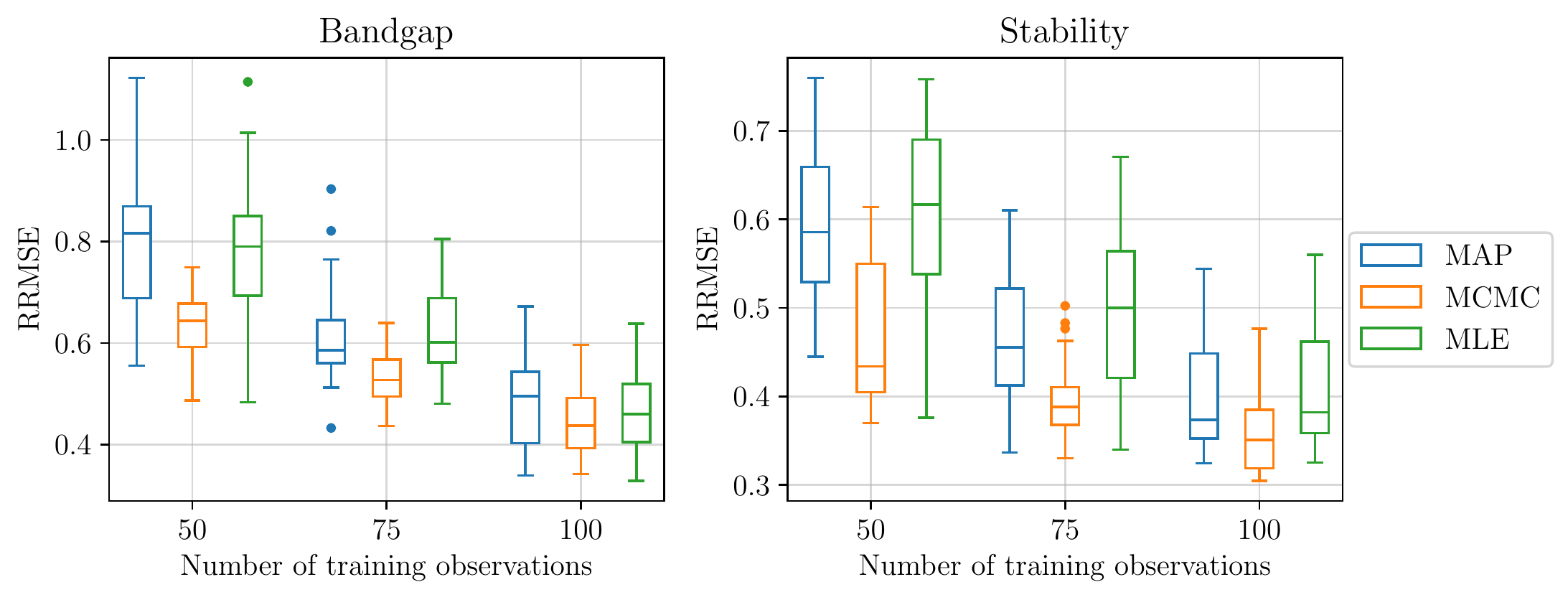}
    \caption{RRMSE results across 25 replicates for the two properties of interest in the lacunar spinel dataset. }
    \label{fig:spinels-rrmse-mle}
\end{figure}

\begin{figure}[H]
    \centering
    \includegraphics[width=0.8\textwidth]{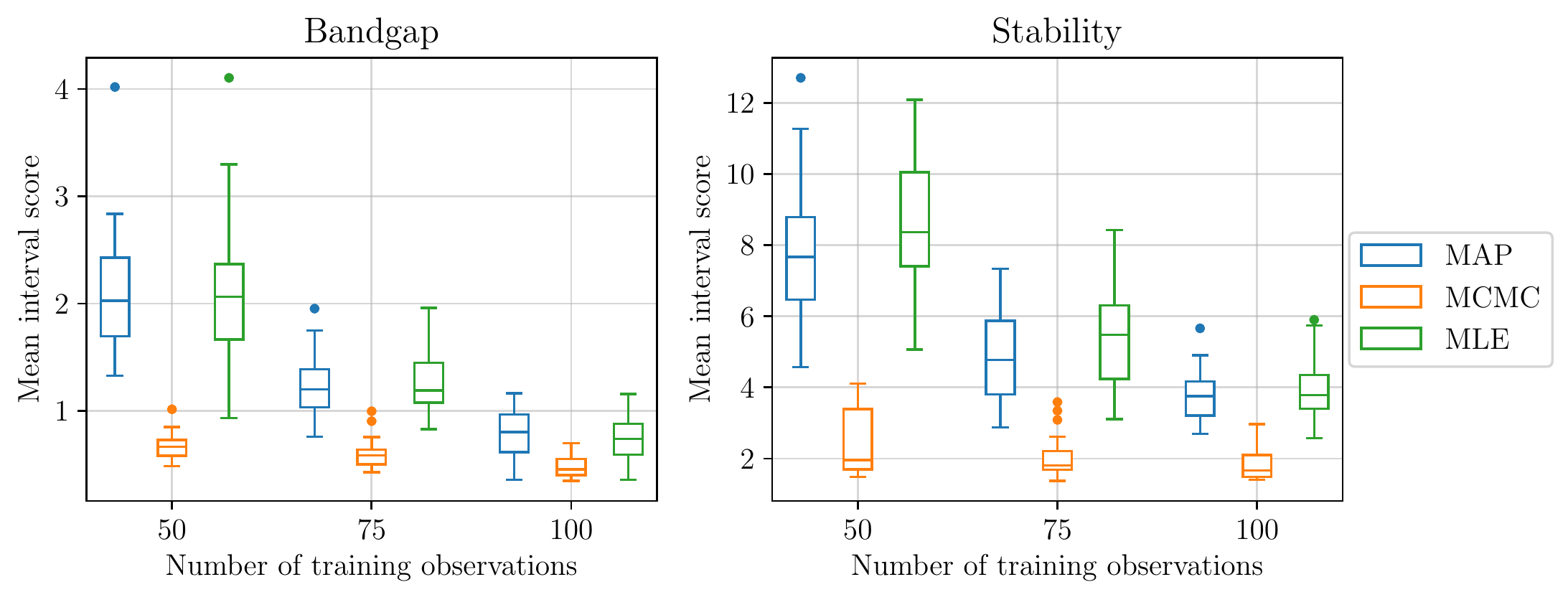}
    \caption{MIS results across 25 replicates for the two properties of interest in the lacunar spinels dataset.}
    \label{fig:spinels-mis-mle}
\end{figure}

\begin{figure}[H]
    \centering
    \subfloat[RRMSE]{
        \label{fig:m2ax-rrmse-mle}
        \includegraphics[width=0.45\textwidth]{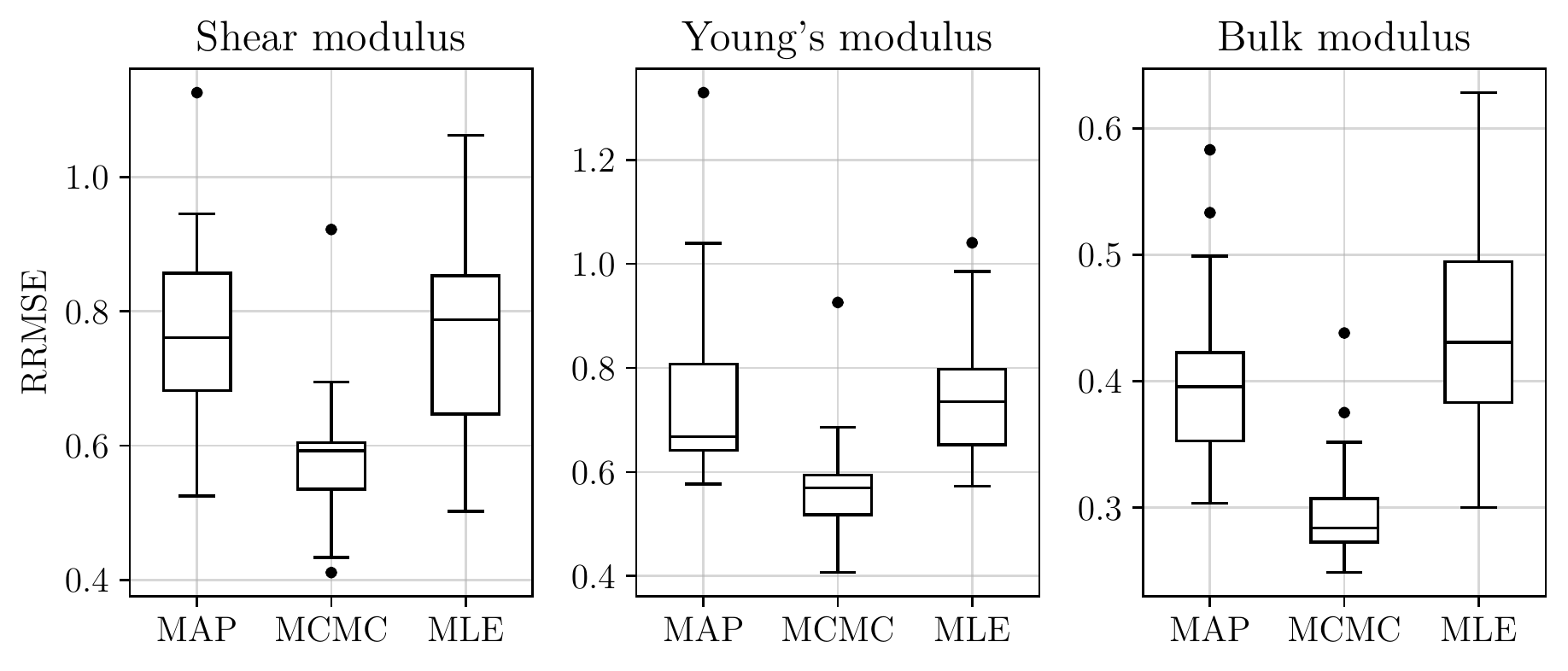}
    }
    \quad
    \subfloat[MIS]{
        \label{fig:m2ax-mis-mle}
        \includegraphics[width=0.45\textwidth]{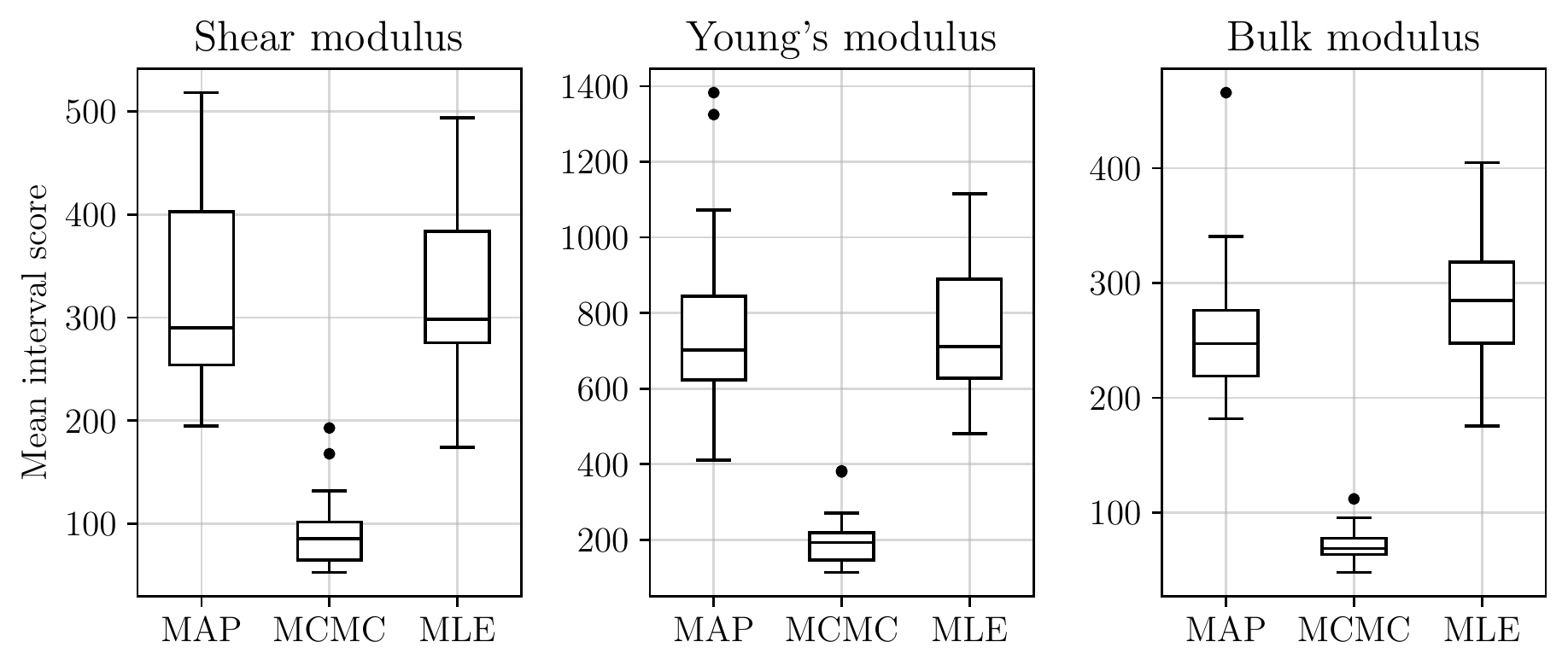}
    }
    \caption{Metrics for the different LVGP models for the three responses in the $M_2AX$ dataset}
    \label{fig:m2ax-mle}
\end{figure}

In the numerical studies, we have used the MAP estimates for plug-in inference. Alternatively, one can use the maximum likelihood estimates (MLEs). The MLE can be viewed as a MAP estimate with a completely non-informative prior. The performance of the MLE estimates for the LVGP models on the examples in Section 5.1 and 5.2 are shown in Figures \ref{fig:math-rrmse-mle} and \ref{fig:math-mis-mle} for the engineering models, Figure \ref{fig:spinels-rrmse-mle} and \ref{fig:spinels-mis-mle} for the lacunar spinels, and Figure \ref{fig:m2ax-mle} for the $M_2AX$ materials. In general, we find that MLE estimates are less  robust and slower to compute than the MAP estimates. 

\section{Optimizing the sparse LVGP approximations}

For the sparse LVGP approximations, one must optimize the regular hyperparameters $\boldsymbol{\theta}$ (including the latent variables), and the parameters for the $M$ inducing points $\mathbf{u}_1,\ldots,\mathbf{u}_M$. For a qualitative variable with $L$ levels, we characterize its inducing points through the weight parameters  described in (4.2), which can be further expressed through $L-1$ bound constrained parameters \cite{betancourt2012cruising}. Thus, there are $M(L-1)$ inducing point parameters for a qualitative variable with $L$ levels. 

Let $\mathbf{K}_{NN}$ denote the $N\times N$ covariance matrix of the function values at the training locations $\mathbf{f}$, $\mathbf{K}_{MM}$  denote the $M\times M$ covariance matrix of the function values at the induncing point locations locations $\mathbf{f}_u$, and $\mathbf{K}_{MN}$ denote the $M\times N$ covariance matrix between $\mathbf{f}$ and $\mathbf{f}_u$. Using the notation of \cite{bauer2016understanding}, the likelihood objective for the FITC \cite{snelson2005sparse} the VFE \cite{titsias2010bayesian} approximation is

\begin{multline}
    \mathcal{L}\smb{\boldsymbol{\theta},\mathbf{u}_1,\ldots,\mathbf{u}_M} = -\frac{N}{2}\log{2\pi} \\  - \underbrace{\frac{1}{2}\log{|\mathbf{G} + \mathbf{Q}_\mathrm{NN}|}}_\text{complexity penalty} - \underbrace{\frac{1}{2}\mathbf{y}^\mathsf{T}\smb{\mathbf{G} + \mathbf{Q}_\mathrm{NN}}^{-1}\mathbf{y}}_\text{data fit} -\frac{1}{2\sigma^2_\epsilon}\mathrm{tr}\smb{\mathbf{T}},
    \label{eq:sparseobj}
\end{multline}

\noindent where $\mathbf{Q}_\mathrm{NN} = \mathbf{K}_\mathrm{MN}^\mathsf{T}\mathbf{K}_\mathrm{MM}^{-1}\mathbf{K}_\mathrm{MN}$ is a rank-M approximation to $\mathbf{K}_\mathrm{NN}$, and 
\begin{align}
    \mathbf{G}_\mathrm{FITC} &= \sigma^2_\epsilon\mathbf{I}_N +  \mathrm{diag}\smb{\mathbf{K}_\mathrm{NN}-\mathbf{Q}_\mathrm{NN}} & \mathbf{G}_\mathrm{VFE} &= \sigma^2_\epsilon\mathbf{I}_N \\
    \mathbf{T}_\mathrm{FITC} &= \mathbf{0} & \mathbf{T}_\mathrm{VFE} &= \mathbf{K}_\mathrm{NN}-\mathbf{Q}_\mathrm{NN}.
\end{align}

\noindent This objective for the sparse LVGP approximations can be optimized with any standard numerical optimization algorithm. In our experiments, we use the L-BFGS-B algorithm \cite{byrd1995limited,zhu1997} with multiple restarts, since the objective \eqref{eq:sparseobj} typically has multiple local optima.  In practice, we find that models with the FITC approximation are much easier to optimize than models with the VFE approximation. We also find that initializing the optimization for the VFE model using the global optima of the corresponding FITC model finds the global optima for the VFE model much faster than with multi-start numerical optimization alone.